\documentclass{article}
\usepackage[preprint]{colm2025_conference}

\usepackage{mathtools}
\usepackage{amsthm}
\usepackage{url}            % simple URL typesetting
\usepackage{booktabs}       % professional-quality tables
\usepackage{amsfonts}       % blackboard math symbols
\usepackage{nicefrac}       % compact symbols for 1/2, etc.
\usepackage{microtype}      % microtypography
\usepackage{graphicx}
\usepackage{amsmath,amssymb} % define this before the line numbering.
\usepackage{color}
\usepackage{epsfig}
\usepackage{tabularx}
\usepackage{multirow}
\usepackage{colortbl}
\usepackage{array}
\usepackage{xspace}
\usepackage{balance}
\usepackage{enumitem}
\usepackage{mathtools}
\usepackage{rotating}
\usepackage{tabulary}
\usepackage[export]{adjustbox}
\usepackage{diagbox}
\usepackage{makecell}
\usepackage{bbding}
\usepackage{pifont}
\usepackage{stfloats}
\usepackage{subfloat}
\usepackage{xspace}
\usepackage{tikz}
\usepackage{bbm}
\usepackage{colortbl,booktabs}
\usepackage{makecell}
\usepackage{diagbox} 
\usepackage{multicol}
\usepackage{array}
\usepackage{amsmath, bm}
\usepackage{amsthm}
\usepackage{dsfont}
\usepackage{subcaption, multirow, overpic, textpos}

\definecolor{green}{RGB}{0,150,10}
\definecolor{blue}{RGB}{0,148,181}
\definecolor{orange}{RGB}{194,153,107}
\usepackage[colorlinks,linkcolor=red,anchorcolor=green,citecolor=blue]{hyperref}

\usepackage[capitalize,noabbrev]{cleveref}

\theoremstyle{plain}

\theoremstyle{definition}

\theoremstyle{remark}

\usepackage[textsize=tiny]{todonotes}
\usepackage{ctable}

\title{ChartX \& ChartVLM: A Versatile Benchmark and Foundation Model for Complicated Chart Reasoning}

\author{Renqiu Xia$^{1,2,*}$, Bo Zhang$^{1,*,\ddagger}$, Hancheng Ye$^{1,*}$,  Xiangchao Yan$^1$, Qi Liu$^{1,2}$, Hongbin Zhou$^1$ \\
\bf{Zijun Chen}$^{1,2}$, \bf{Peng Ye}$^{1}$, \bf{Min Dou}$^1$, \bf{Botian Shi}$^{1}$, \bf{Junchi Yan}$^{2,\ddagger}$, \bf{Yu Qiao}$^1$ \\[2mm]
$^1$Shanghai Artificial Intelligence Laboratory  \\ $^2$ Shanghai Jiao Tong University \\
}

\begin{document}

\newcommand\blfootnote[1]{%
\begingroup
\renewcommand\thefootnote{}\footnote{#1}%
% \addtocounter{footnote}{-1}%
\endgroup
}

\blfootnote{* Core Contributor, $\ddagger$ Corresponding Authors \\}

\maketitle

\begin{abstract}

Recently, many versatile Multi-modal Large Language Models (MLLMs) have emerged continuously. However, their capacity to query information depicted in visual charts and engage in reasoning based on the queried contents remains under-explored. In this paper, to comprehensively and rigorously benchmark the ability of the off-the-shelf MLLMs in the chart domain, we construct ChartX, a multi-modal evaluation set covering 18 chart types, 7 chart tasks, 22 disciplinary topics, and high-quality chart data. Besides, we develop ChartVLM to offer a new perspective on handling multi-modal tasks that strongly depend on interpretable patterns, such as reasoning tasks in the field of charts or geometric images. We evaluate the chart-related ability of mainstream MLLMs and our ChartVLM on the proposed ChartX evaluation set. Extensive experiments demonstrate that ChartVLM surpasses both versatile and chart-related large models, achieving results comparable to GPT-4V. We believe that our study can pave the way for further exploration in creating a more comprehensive chart evaluation set and developing more interpretable multi-modal models. Both ChartX and ChartVLM are available at: \textcolor{teal}{\url{https://github.com/Alpha-Innovator/ChartVLM}}

\end{abstract}

% \vspace{-5pt}
\section{Introduction}
\vspace{-3pt}

Versatile Multi-modal Large Language Models (MLLMs)~\citep{bai2023qwen, 2023gpt4V, achiam2023gpt, chen2024far, 2024claude, xia2024docgenome, reid2024gemini} have made promising progress in general-purpose vision-language applications such as multi-modal Question Answering (QA)~\citep{lin2023sphinx, bai2023qwen, wang2023cogvlm, liu2023improved}, embodied AI~\citep{huang2023instruct2act}, and mathematical reasoning~\citep{trinh2024solving, yang2023leandojo, jiang2022draft}. Although MLLMs have demonstrated their powerful generalization ability in a wide range of multi-modal tasks, their performance in multi-modal reasoning tasks still falls short of human abilities~\citep{yang2023dawn, bubeck2023sparks, wang2024cdm, achiam2023gpt}. For instance, humans can easily extract numerical values from a given visual chart and engage in a series of complicated logical reasoning based on the extracted values. However, at present, the MLLMs' ability to perform complicated logical reasoning based on chart data has not been fully explored.

In this paper, to further validate their capabilities in more complicated reasoning tasks involving chart data, we propose a multi-modal benchmark for comprehensive chart understanding. As illustrated in Fig.~\ref{fig1:motivation}, our work comprises two contributions: \textbf{(1)} ChartX, which is a comprehensive, high-quality evaluation set designed to sufficiently assess the chart understanding abilities of the off-the-shelf MLLMs, and \textbf{(2)} An interpretable Chart-domain Vision-Language Model (ChartVLM) for general-purpose chart applications.

To construct a comprehensive chart evaluation set, we collected 48K multi-modal chart data covering 22 topics, 18 chart types, and 7 tasks. Each chart data within this dataset includes four modalities, including image, Comma-Separated Values (CSV), python code, and text description. According to the task complexity, we classify the proposed 7 chart tasks into two general categories: perception tasks (chart structural extraction, chart type classification, and chart title extraction) and cognition tasks (QA, chart description, chart summarization, and chart re-drawing).

% When conducting multimodal reasoning tasks in certain scientific domains that rely heavily on interpretability, our primary observation is the execution of perception tasks before engaging in reasoning tasks. The statistical information extracted during the perception tasks plays a crucial role in providing interpretability for the model's reasoning tasks. Building upon this, we introduce ChartVLM, characterized by reasoning tasks referencing the predictive results of perception tasks to enhance the reliability of the reasoning process. Additionally, ChartVLM employs an instruction adapter to dynamically select tasks based on user instructions, ensuring both interpretability and interactivity.
    
For certain scientific domains such as chart reasoning, where interpretability is paramount, our primary observation is prioritizing the perception tasks before engaging in the complicated reasoning tasks. The statistical information extracted via the perception tasks plays a pivotal role in providing essential support for the interpretability of the model's reasoning tasks. Building upon this observation, we introduce ChartVLM, characterized by the integration of perception task predictions (\textit{e.g.}, structural data extraction) into reasoning tasks
% the fact that reasoning tasks incorporate the prediction results of perception tasks (\textit{e.g.} structural data extraction)
to enhance the interpretability of the reasoning results. Furthermore, ChartVLM utilizes an instruction adapter to dynamically select tasks that users expect to perform according to the users' instructions, ensuring both interpretability and interactivity.

On top of this, the existing open-source chart datasets are consolidated for the training of ChartVLM, including ChartQA~\citep{masry2022chartqa}, Chart-to-text~\citep{obeid2020chart}, PlotQA~\citep{methani2020plotqa}, and SimChart9K~\citep{xia2023structchart}.
% , and train the proposed ChartVLM on the merged dataset. 
Note that during the training process, ChartVLM has no access to any data from the ChartX evaluation set. Then, we conduct comprehensive comparisons of ChartVLM with current MLLMs~\citep{bai2023qwen, liu2023improved} on the ChartX evaluation set, including base abilities, \textit{e.g.}, data extraction, and advanced abilities, \textit{e.g.}, complicated problem-solving, where we demonstrate the superiority of our ChartVLM.
% \vspace{-2pt}
\section{Related Work}

\vspace{-5pt}
\begin{figure*}[t]
    \centering
    \resizebox{0.99\linewidth}{!}{\includegraphics{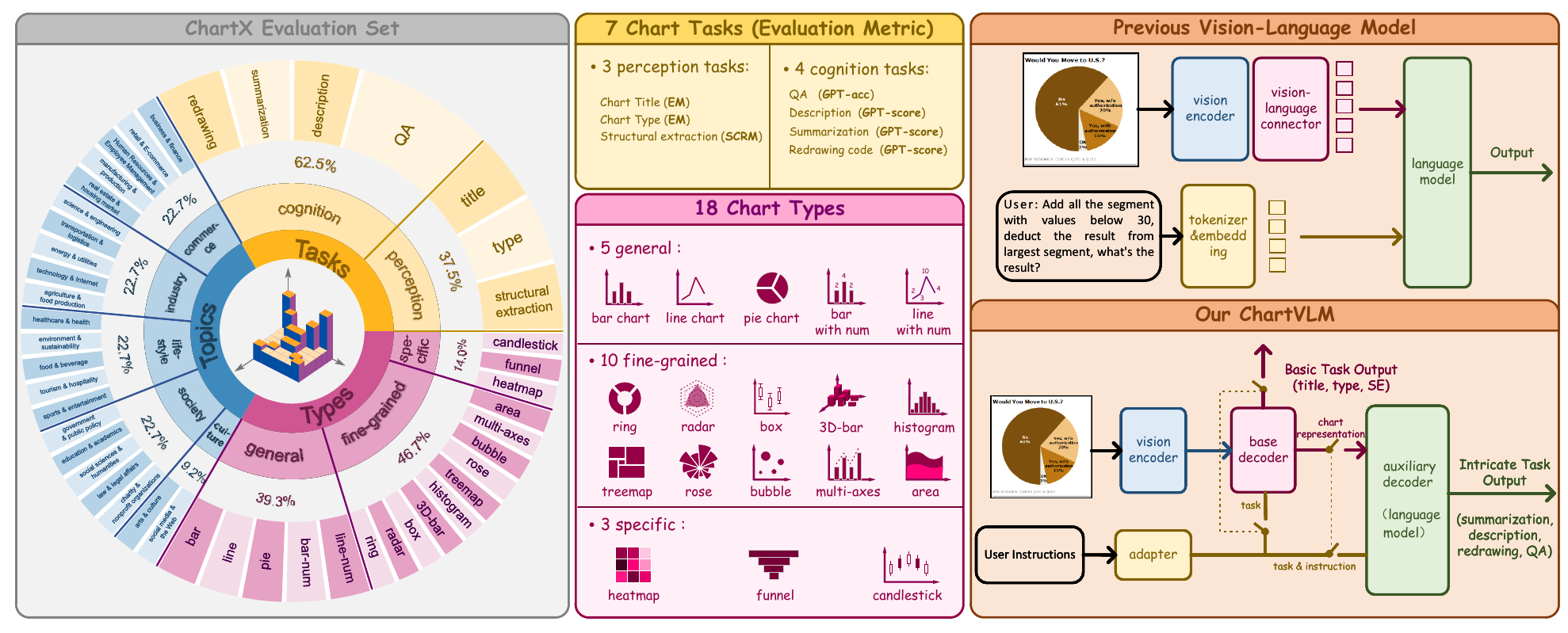}}
    \vspace{-6pt}
    \caption{Our work offers two insights: \textbf{a)} ChartX: a comprehensive multi-modal chart evaluation set encompassing 22 disciplinary topics, 18 chart types, and 7 tasks where models are evaluated using task-specific metrics such as EM, GPT-acc GPT-score, SCRM~\citep{xia2023structchart}, and \textbf{b)} ChartVLM: a novel framework to perform the multi-tasks in the chart domain. Our key point is to leverage the instruction adapter to dynamically choose the task that needs to be executed. For downstream tasks that rely on querying chart information, we prioritize chart structural extraction before engaging in chart reasoning tasks. This sequence aims to enhance the interpretability of the reasoning results.} 
    \label{fig1:motivation}
\vspace{-2pt}
\end{figure*}

\begin{table*}[t]
\vspace{-4pt}
\centering
\small
\caption{Comparison with the existing chart-related benchmarks, where ChartX is constructed for comprehensively evaluating the off-the-shelf vision-language large models from more chart types and topics. Besides, EM denotes Exact Match and SCRM represents the Structuring Chart-oriented Representation Metric described in StructChart~\citep{xia2023structchart}.}

\resizebox{0.99\textwidth}{!}
{
\begin{tabular}{lccccccc@{}}
\toprule
\multirow{2}{*}{Study Works}           & \multirow{1}{*}{\# Chart}            & \# Chart & \# Task & \# Evaluation & \# Evaluation  & Evaluation &Open- \\ 
  & Topic & Type & Type & Chart Images & Dataset & Metric &source\\ 
\midrule
\textit{~~Single-task Evaluation} \\
PlotQA~\citep{methani2020plotqa}          & N/A               & 3   & 1                  & 33.7K           &  PlotQA           &       EM \& AP    & \Checkmark                 \\
Chart-to-text~\citep{obeid2020chart}        & 6          & 6       & 1         & 6.6K            & Chart-to-text                    &       EM          & \Checkmark                 \\
ChartQA~\citep{masry2022chartqa}   & 15   & 3         & 1    & 1.5K                & ChartQA             &       EM          &  \Checkmark               \\
OpenCQA~\citep{kantharaj2022opencqa}    & 10                     & 5     & 1             & 1.2K & OpenCQA              &    EM \& BLEU \& ROUGE       & \Checkmark                    \\ 
\midrule
\textit{~~Multi-task Evaluation} \\
ChartLlama~\citep{han2023chartllama}        & N/A              & 10      & 7          & 1.5K            & ChartQA \& Chart-to-text              &       EM \& GPT           & \XSolidBrush             \\
ChartBench~\citep{xu2023chartbench}      & N/A             & 9       & 4          & 2K             & ChartBench                  &       Accuracy          & \XSolidBrush                 \\
MMC~\citep{liu2023mmc}          & 5                    & 6         & \textbf{9}        & 2.1K        & MMC          &       GPT     & \XSolidBrush               \\
ChartAssisstant~\citep{meng2024chartassisstant}          & N/A                    &     9     &  5       & 1.5K         & ChartQA \& OpenCQA   &       EM\& BLEU           & \XSolidBrush               \\
\textbf{Ours}     & \textbf{22}   & \textbf{18} & 7 & \textbf{6K} &  \textbf{ChartX}  & \textbf{EM} \& \textbf{SCRM} \& \textbf{GPT-acc}  \& \textbf{GPT-score} & \Checkmark  \\
\bottomrule
\end{tabular}}

\label{tab:related}
\end{table*}

\textbf{Chart Perception} aims to extract the numerical and textual information from a given visual chart. By leveraging the OCR tools~\citep{luo2021chartocr} to supplement the textual information, the basic function of extracting chart information can be achieved. Recently, some researchers~\citep{hassan2023lineex, rane2021chartreader, huang2023lvlms} have attempted to perform a chart-to-table transformation for the visual chart perception task, by means of self-supervision from image-table pairs. For example, Deplot~\citep{liu2022deplot} fine-tuned an image-to-text transformer for such conversion. StructChart~\citep{xia2023structchart} utilizes the encoder-decoder framework to achieve transformation. These methods extract the tabular format of a visual chart and leverage the external module such as GPT~\citep{ouyang2022training, brown2020language} to perform downstream tasks. However, their chart-related reasoning abilities strongly depend on external modules, whose scalability is hard to guarantee.

\textbf{Chart Cognition} is defined as a process to deal with intricate tasks related to both chart-related knowledge and common sense knowledge.
A typical example is to query numerical points from a chart and give the prediction results using mathematical or logical reasoning. Recent studies~\citep{he2023text2analysis, tian2023chartgpt, zha2023tablegpt, lee2023pix2struct, liu2022matcha} focus on showing the reasoning ability of their models on chart domain. Pix2Struct~\citep{lee2023pix2struct} presents a pre-training method using masked screenshots from web pages, which is verified to be effective in chart understanding tasks such as ChartQA dataset~\citep{masry2022chartqa}. Besides, MatCha~\citep{masry2022chartqa} decodes the answers to chart questions in an end-to-end manner, where the chart reasoning ability can be enhanced from MATH data~\citep{saxton2019analysing}.  

% to enhance the model reasoning ability from chart domain
\textbf{Multi-Modal Chart Generation and Benchmark.} 
Chart data generation is a crucial step for scaling up the model ability~\citep{tian2023chartgpt, liu2022matcha, akhtar2023chartcheck}. Previous chart-related benchmarks only cover general three types of charts (line, pie, bar charts) and focus on a few tasks such as chart-to-table tasks for ChartQA~\citep{masry2022chartqa}, PlotQA~\citep{methani2020plotqa}, and Chart-to-Text~\citep{obeid2020chart}, and QA tasks for DVQA~\citep{kafle2018dvqa} and OpenCQA~\citep{kantharaj2022opencqa}. Recently, various benchmarks
% , such as MMC~\citep{liu2023mmc}, ChartLlama~\citep{han2023chartllama}, ChartBench~\citep{xu2023chartbench}, and ChartAssisstant~\citep{meng2024chartassisstant}, 
have been proposed in some works, \textit{e.g.} MMC~\citep{liu2023mmc}, ChartLlama~\citep{han2023chartllama}, ChartBench~\citep{xu2023chartbench}, and ChartAssisstant~\citep{meng2024chartassisstant}, with the common characteristics of more types, more tasks, and more modalities of chart data, which is insightful for the chart community. However, as shown in Table~\ref{tab:related}, the data and metric diversity of charts used for evaluating multi-modal large models is relatively limited. For example, ChartBench~\citep{xu2023chartbench} {merely uses a two-sided judgment (yes or no) to evaluate model performance. The types of charts and data in MMC~\citep{liu2023mmc} are also insufficient in verifying the chart ability of the off-the-shelf MLLMs.

% For example, the test set of MMC only contains 854 images and 3 general chart types. By comparison, we are committed to building a chart evaluation benchmark with multiple types, tasks, and modalities to comprehensively demonstrate the performance of MMLMs in the chart domain.

% \vspace{-2pt}
\section{ChartX: Multi-task Chart Evaluation Set}
\vspace{-2pt}

\subsection{Coverage Analysis of the Evaluation Set}
\vspace{-2pt}
We describe the coverage range of ChartX from chart types, chart topics, and chart-related tasks, respectively.

\noindent\textbf{Chart Types.} ChartX covers all chart types where chart data can be directly converted into a structural data format, \textit{e.g.}, CSV format, resulting in a total of more than 18 chart types. For a clear visualization, we categorize different chart types into three groups based on their usage frequency and application fields. \textbf{(1) General Chart Types}: bar chart (with or w/o numerical data), line chart (with or w/o numerical data), and pie chart. These five chart types are commonly employed to represent a wide range of chart data distribution. \textbf{(2) Fine-grained Chart Types}: ring chart, radar chart, box plot, 3D-bar chart, histogram, treemap, rose chart, bubble chart, multi-axes chart, and area chart. These 10 chart types are mostly variations of the general chart types to present the complex data distribution more vividly. \textbf{(3) Domain-specific Chart Types}: heatmap, funnel, and candlestick. These three chart types are specially designed to visualize data distribution within domain-specific fields. For example, heatmap is commonly used to visualize the significant difference trend in a 2D space. Funnel charts are widely used in the analysis of market sales, while candlestick is primarily utilized for depicting stock trends. The distribution statistics of chart type in ChartX are shown in Fig.~\ref{fig1:motivation}. Specifically, we generate more images on general chart types to expand the chart diversity, which are more frequently utilized with more diversity. For the fine-grained chart types, the image number of each type is balanced to avoid the long-tail distribution issue in our benchmark.

\noindent\textbf{Chart Topics.} ChartX contains various chart topics covering as many themes as possible. Specifically, the high-level topics in ChartX can be divided into five perspectives: commerce, industry, society, culture, and lifestyle. And fine-grained topic types can be subdivided into 22 sub-disciplinaries, which are listed in Fig.~\ref{fig1:motivation}. The topic distribution of ChartX is presented in Fig.~\ref{fig1:motivation}. More statistical results of chart topics are shown in Appendix~\ref{sec:chart_topic}.

\noindent\textbf{Chart Tasks.} Unlike previous chart benchmarks focusing on the category of visual logic reasoning tasks, the ChartX benchmark emphasizes the interpretability for all downstream chart-related tasks. Given that interpretability relies heavily on the ability to perceive chart information, ChartX categorizes perception-related tasks as base tasks, including title perception, chart type recognition, and Structural Extraction (SE). On the other hand, other chart-related tasks are classified as intricate cognition tasks, including chart-related Question Answering (QA), Chart Description, Chart Summarization, and Chart Redrawing. In the context of ChartX, \textbf{QA} refers to answering questions that are formulated solely based on the chart data, requiring reasoning derived directly from the provided chart information. This characteristic distinguishes ChartX from previous chart-related QA datasets like ChartQA~\citep{masry2022chartqa}. In ChartQA~\citep{masry2022chartqa}, there exists a certain number of QA pairs that cannot be answered solely based on the information presented in the given chart image.
%where the questions are proposed from contextual texts that cannot be reasoned from chart images. 
\textbf{Chart Description} aims at presenting detailed information and some insights from the distribution of chart data, while \textbf{Chart Summarization} features summarizing the trend-like or high-level characteristics from the given data in a few sentences. \textbf{Chart Redrawing} refers to plotting the given data into a new chart image with the same chart type of original data. The distribution of each task is listed in Fig.~\ref{fig1:motivation}. For each image, together with labels of base tasks, we collect two QA samples, one description sample, one summarization sample, and one redrawing code sample. Overall, the samples from multi-tasks reach 48K in ChartX.

% \vspace{-3pt}
\subsection{Distribution Analysis of the Evaluation Set}
\vspace{-3pt}
We analyze the distribution diversity of the ChartX benchmark by considering both style distribution and content distribution. Fig.~\ref{fig:dis_vis} visually depicts the diversity comparison among various chart benchmarks using t-SNE.
%The visual comparison of the benchmark diversity between different chart benchmarks is presented in Fig. \ref{} by using t-SNE \cite{tsne}.

\noindent\textbf{Style Distribution.} In terms of style distribution, the inner-class diversity is considered to augment the style fashion of each chart type. Such diversification is achieved by both package and hyper-parameter diversity performed by human efforts. For each chart type, we design an individual diversification scheme with different plotting package candidates and different hyper-parameter settings. A general alternative plotting scheme includes \textit{matplotlib}, \textit{seaborn}, and \textit{plotly} packages, \textit{etc}, while some domain-specific packages like \textit{mplfinance} are also employed to increase the diversity. The hyper-parameter diversity involves the adjustment of all possible hyper-parameter settings in plotting, \textit{e.g.}, figure size, background setting, axis/legend location, line, marker style, tick, filling styles, alpha, annotation, \textit{etc}.

\noindent\textbf{Content Distribution.}
As for content diversity, the CSV data length distribution and task-wise token distribution for each chart are visualized for different chart benchmarks to compare the content distribution diversity. As shown in Fig.~\ref{fig:dis_vis}, the ChartX benchmark presents a higher diversity in both CSV data length and token distribution than the existing benchmarks.

% \vspace{-3pt}
\subsection{Two-stage Chart Data Generation}
\vspace{-3pt}

\begin{figure}[t]
% \vspace{-2pt}
% \vspace{-2em}
    \centering
    \small
    \subfloat[Image Distribution.]{
    \begin{minipage}{0.24\linewidth}{\begin{center}
    \resizebox{\linewidth}{!}{\includegraphics{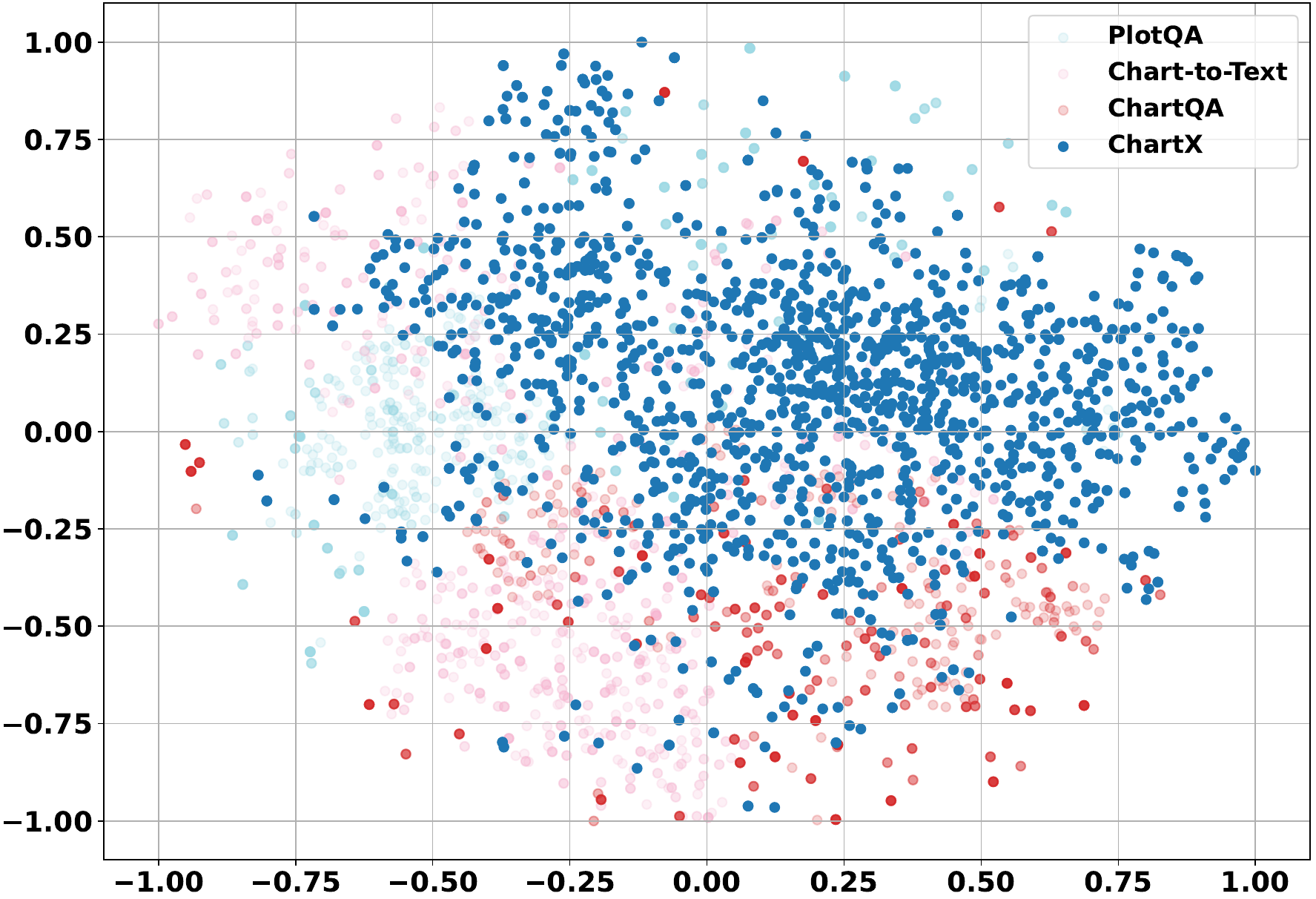}}\end{center}}\end{minipage}
    % \vspace{-10pt}
    \label{fig:exp_se}}
    \subfloat[CSV Data Distribution.]{
    \begin{minipage}{0.24\linewidth}{\begin{center}
    \resizebox{\linewidth}{!}{\includegraphics{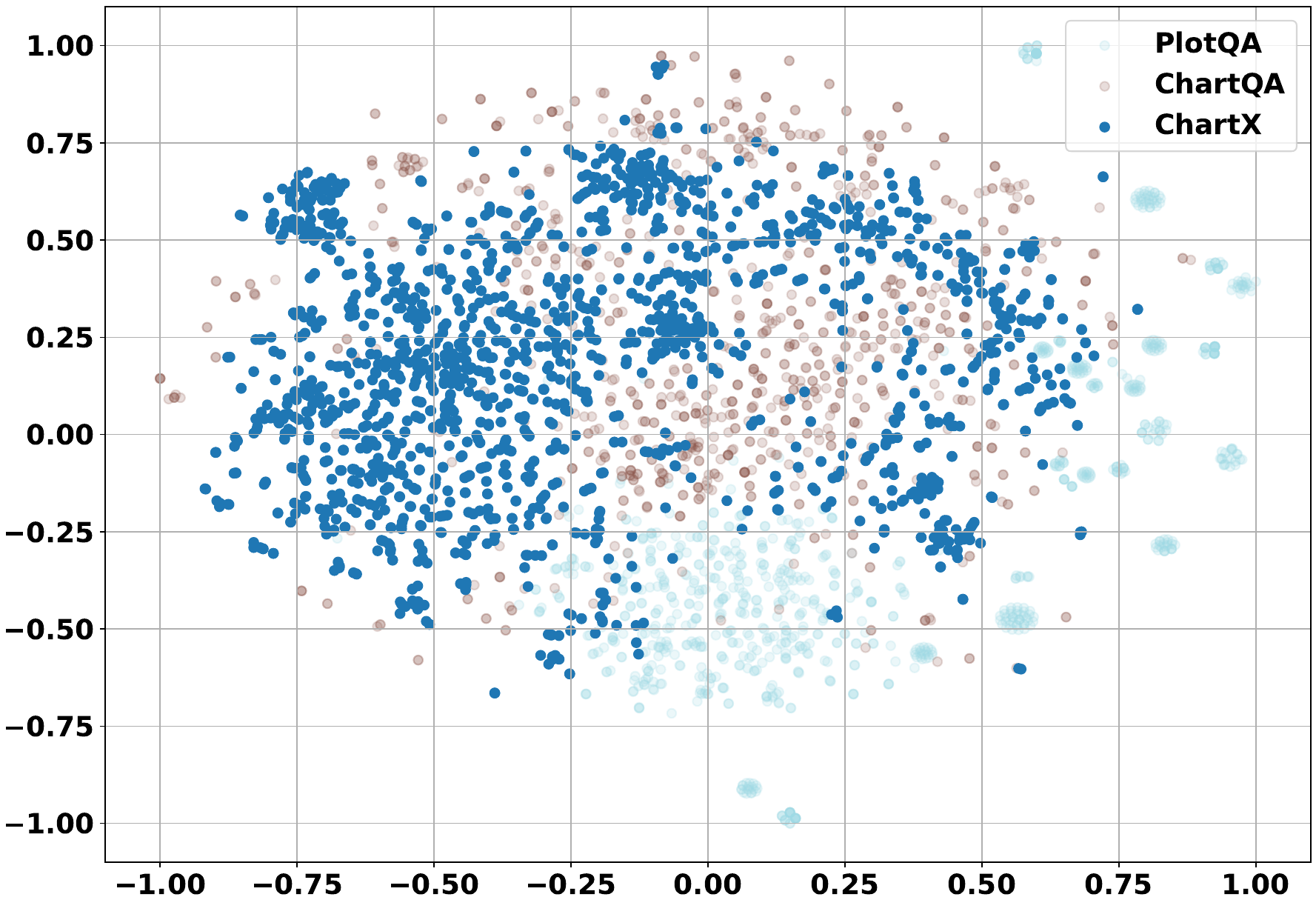}}\end{center}}\end{minipage}
    \label{fig:exp_qa}}
    \subfloat[Question Distribution.]{
    % \resizebox{\linewidth}{!}{
    \begin{minipage}{0.24\linewidth}{\begin{center}
    \resizebox{\linewidth}{!}{\includegraphics{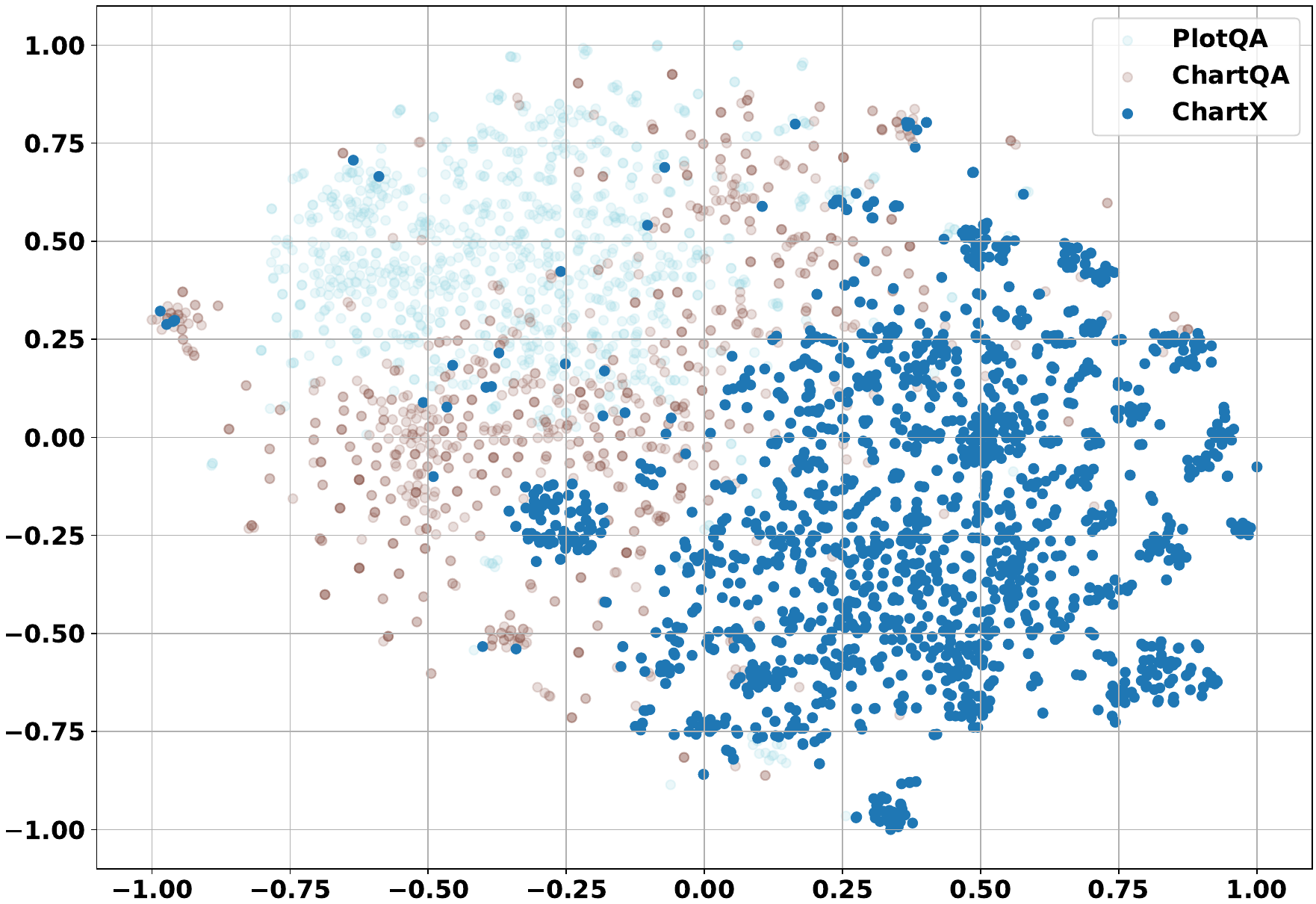}}\end{center}}\end{minipage}
    % \vspace{10pt}
    \label{fig:exp_summ}
    }
    \subfloat[Data Length Distribution.]{
    % \resizebox{\linewidth}{!}{
    \begin{minipage}{0.24\linewidth}{\begin{center}
    \resizebox{\linewidth}{!}{\includegraphics{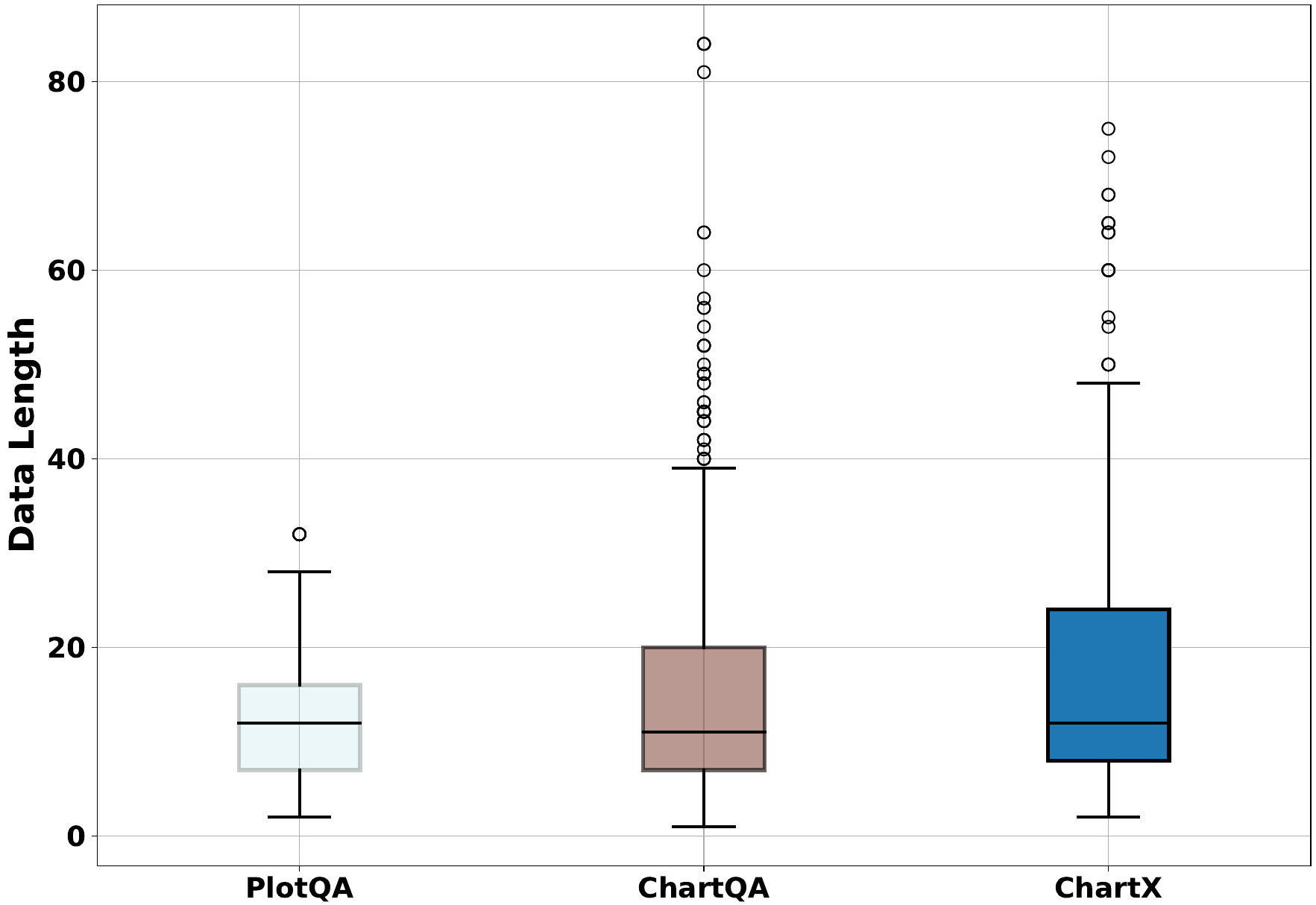}}\end{center}}\end{minipage}
    % \vspace{10pt}
    \label{fig:exp_desc}
    }
    \vspace{-4pt}
    \caption{Data Distribution Comparisons, depicting the diversity of (a) chart image, (b) CSV data, (c) questions in QA pairs, and (d) CSV data length.}
    \label{fig:dis_vis}
\vspace{-4pt}
\end{figure}
% (a) indicates more image diversity of ChartX, (b) shows the diversity of the generated CSV data, (c) presents the diversity of instruction data, and (d) depicts the wider distribution of CSV numerical data length than existing evaluation benchmarks.

Utilizing the strong generation capabilities of GPT-4~\citep{achiam2023gpt}, ChartX is created through an automated online generation process with manual instructions. This involves a data-centric two-stage generation paradigm, encompassing the creation of perception and cognition data.

% As described above, the perception data in chart images consist of chart data, chart title, and chart type. For the generation of chart titles and chart types, we first define the selection space of chart topics and chart types, where both spaces are first initialized by GPT-4 and then adjusted by humans to closely resemble the real-world chart contents and match the practical potential of conversion to CSV-format data for the selected chart types. For the generation of chart data, we utilize GPT-4 to generate the real data distribution with specific requirements of data length for the provided chart type and chart topic.

\noindent\textbf{Data Acquisition: Chart Perception.} As mentioned earlier, chart perception data includes chart data, chart title, and chart type. To generate chart titles and types, we initialize selection spaces with GPT-4, which are later refined by human adjustment (refer to Appendix~\ref{para:validation}) to align closely with real-world chart contents and ensure practical conversion potential to CSV-format data. For chart data generation, GPT-4 is employed to generate the actual data distribution based on the specified length requirements for the given chart type and chart topic.

\begin{figure*}[!t]
\vspace{-4pt}
    \centering
    \resizebox{0.99\linewidth}{!}{\includegraphics{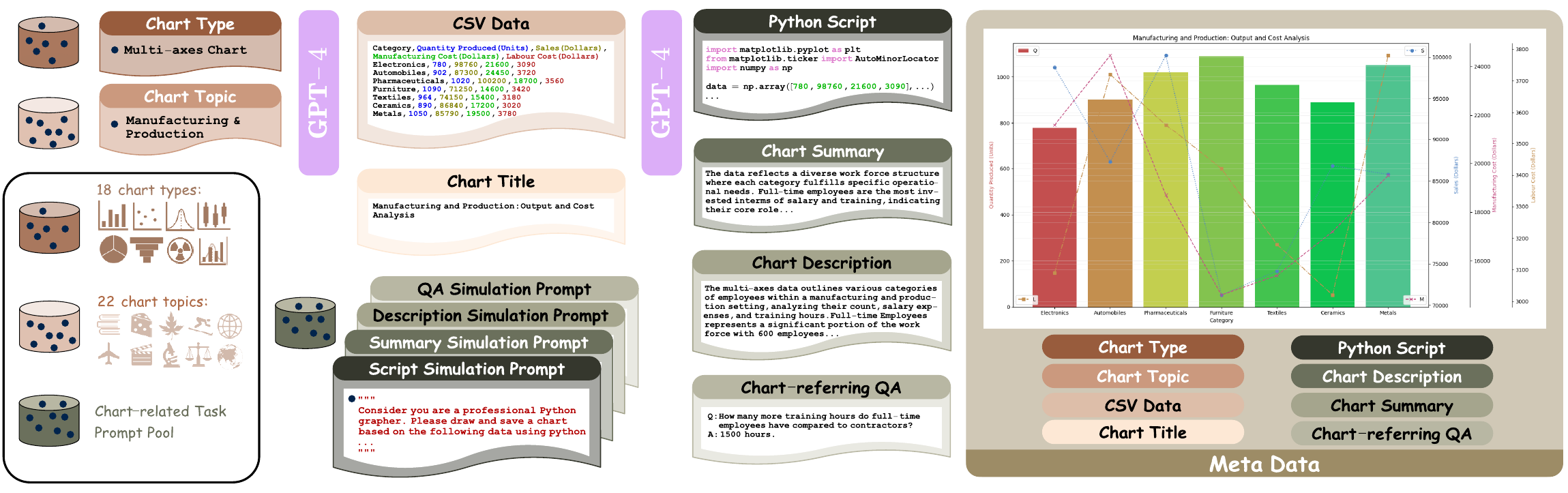}}
    % \vspace{-8pt}
    \caption{Pipeline of chart data acquisition followed by a manual quality inspection process as introduced in Appendix~\ref{para:validation}. For different chart tasks, we design different prompts and data generation processes around 22 chart topics and 18 chart types to enhance the data diversity in the chart domain.} 
    \label{fig:generation}
    \vspace{-2pt}
\end{figure*}

\noindent\textbf{Data Acquisition: Chart Cognition.} The generation of chart cognition data is based on the generated chart perception data. For each chart perception data sample, we design individual instructions with special task templates (refer to Appendix~\ref{sec:specific_type_gen}) to generate different cognition task data. Besides, some chart type-specific instruction examples will be randomly sampled to guide the data generation, which is widely and specially designed for the corresponding chart type and topic. Among these tasks, the generated redrawing code is utilized to render the chart image, constructing the image-label pairs as metadata for the ChartX benchmark, which is further illustrated in Fig.~\ref{fig:generation} and Appendix~\ref{sec:data_gen}.

% \vspace{-5pt}
\subsection{Task Evaluation Metrics}
\label{sec:eval}
\vspace{-3pt}

\noindent\textbf{SCRM.} Given that data in the chart has matrix-like row-column transformation invariance and transpose transformation invariance, Structuring Chart-oriented Representation Metric (SCRM)~\cite{xia2023structchart} is employed to evaluate the extracted chart information (\textit{i.e.} SE task), in which the linearized CSV tokens predicted by all models will be transformed to triplet format for performing SCRM evaluation. 
% In this work, we consistently employ the SCRM as established in StructChart~\cite{xia2023structchart} to evaluate the performance of SE task.

% Due to the inherent variability of specific data classifications
% This enables a refined assessment of the model’s precision and facilitates a more sophisticated interpretation of quantitative information.
\noindent\textbf{GPT-acc \& GPT-score.} The GPT-acc metric is designed for tasks with unambiguous answers like question-answering, where outputs are evaluated against an exact ground truth using GPT-4. To make a rational evaluation, GPT-acc incorporates a 5\% margin of error for numerical responses. Conversely, the GPT-score metric addresses open-ended tasks where responses are subjectively graded. Here, GPT-4 rates summarization, description, and code-redrawing outputs on a \textbf{0-5} scale based on manually adjudicated scoring criteria. All the prompts about the manual criteria for each task are described in Appendix~\ref{sec:eval_setting}, which considers completeness, relevance, accuracy, and creativity of responses.

\begin{figure*}[tb!]
\vspace{-2pt}
    \centering
    \includegraphics[width=0.98\textwidth]{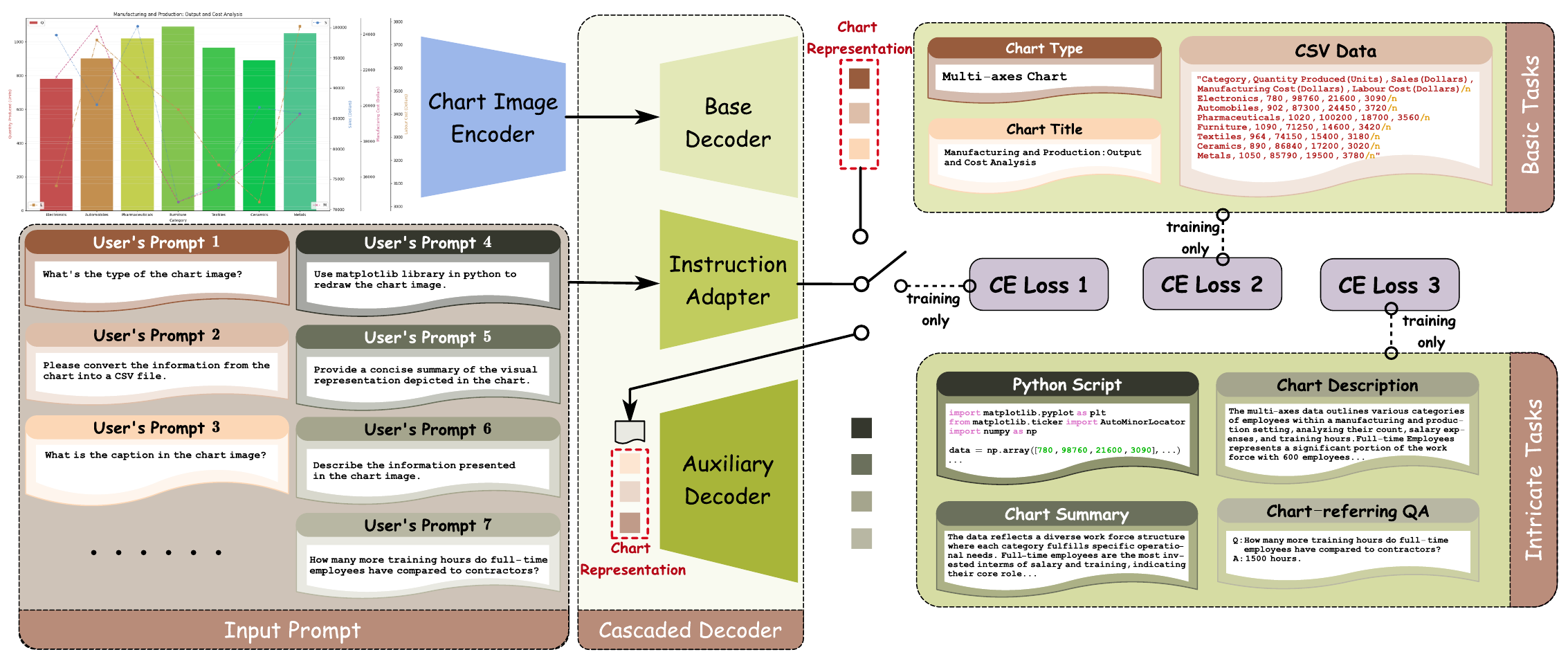}
    \vspace{-4pt}    
    \caption{ChartVLM Overview: \textbf{a)} To enhance the interpretability of the chart model in cognition tasks (\textit{e.g.} answer questions based on chart image), ChartVLM first performs the base perception task (\textit{e.g.} structural extraction from the given chart image to a predicted CSV data), and then, finishes other cognition tasks (\textit{e.g.} chart redrawing, description, summary, and QA) based on the extracted structural data. \textbf{b)} To choose the task that users expect to perform according to the prompts they use, the instruction adapter is designed to cover a variety of user instructions as illustrated in this figure.} 
    \label{fig:framework}
    \vspace{-4pt}
\end{figure*}

% \vspace{-2pt}
\section{ChartVLM: Chart Vision-Language Model}

% \vspace{-2pt}
\subsection{Overall Model Design}
\vspace{-3pt}

Here, we introduce ChartVLM, an innovative framework illustrated in Fig.~\ref{fig:framework}. This architecture comprises an instruction adapter, a pixel-level encoder, and a pair of text-level cascaded decoders. The instruction adapter serves as the initial chart task routing module, selecting chart tasks to be executed based on the user's instructions. For base tasks, such as the prediction of chart title, type, and CSV data, only the base decoder engages. Conversely, the auxiliary decoder will be activated for more intricate generative tasks, building upon the CSV predictions obtained by the base decoder. 

\vspace{-3pt}
The motivations of the cascaded mechanism are: 1) to \textbf{augment the model's interpretability} in cognition tasks through the incorporation of intermediate chart representations, such as CSV data and title, type, and \textit{etc}, and 2) to \textbf{improve computational efficiency} by allocating the workload across decoders of varying parameters, wherein the base decoder is significantly smaller than auxiliary decoder.

\begin{figure*}[!t]
\vspace{-4pt}
    \centering
    \resizebox{1.0\linewidth}{!}{\includegraphics{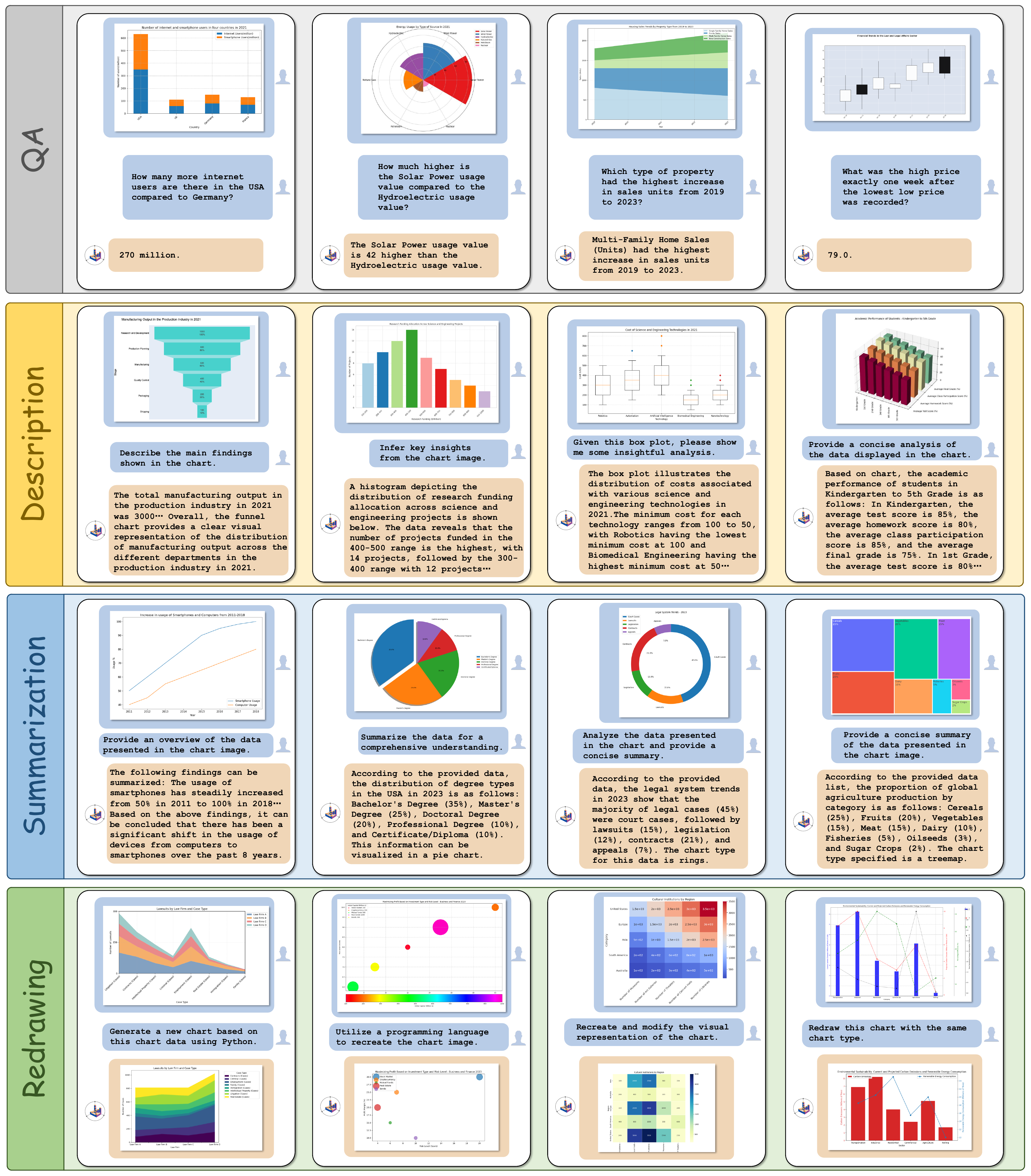}}
    \vspace{-9pt}
    \caption{ChartX visualization results of zero-shot chart images using our ChartVLM model. Here we show 4 cognition tasks and please refer to Appendix~\ref{sec:vis_result_perception_tasks} for more results of perception tasks.} 
    \label{fig6:demo}
\vspace{-4pt}
\end{figure*}

% \vspace{-3pt}
\subsection{Instruction Adapter: Instruction Selection}
\vspace{-3pt}

% the purpose and structure of the proposed instruction-language adapter.
The purposes of designing an instruction adapter are: 1) to meet a broad spectrum of user instructions, and 2) to dynamically select the decoder assigned based on user instructions. The instruction adapter has a simple structure, consisting of only three linear layers, efficiently mapping diverse user instructions to one of seven chart task categories. For training the instruction adapter, we construct a simple dataset using GPT-3.5, containing 7K pairs of user instructions and their task labels. The designed instruction adapter demonstrates flawless performance on the validation subset we constructed, with a 100\% accuracy rate.

% \vspace{-3pt}
\subsection{Cascaded Decoders Design}
\vspace{-3pt}

The base decoder is developed to extract chart information (mainly CSV data) from a visual chart. If a task is classified as a basic perception task by instruction adapter, the chart at pixel-level will be converted to textual representations output directly (\textit{e.g.} chart title, type, and CSV data) \textbf{without the need for auxiliary decoder intervention}. Conversely, when dealing with complicated tasks that require intricate generative processes, the auxiliary decoder will be activated. It leverages both the textual representational outputs from the base decoder and user instructions to execute its sophisticated operations. Once the chart task is determined by the adapter, the cascaded decoders are dynamically and efficiently allocated to meet the varying task requirements.

For basic perception tasks, we fine-tune all the network weights pre-trained from Pix2Struct-base and Pix2Struct-large~\citep{lee2023pix2struct} model, using image-CSV pair data. The fine-tuned encoder and decoder are regarded as chart image encoder and base decoder in ChartVLM. After the fine-tuning stage is completed, the encoder-decoder can effectively transform the chart in image format into a CSV format (\textit{i.e.} chart representation in Fig.~\ref{fig:framework}). For intricate cognition tasks, we utilize LoRA~\citep{lora} and fine-tune the pre-trained Vicuna-7B and Vicuna-13B as auxiliary decoders using text-text pair data including CSV, QA, summarization, and drawing codes. 

Ultimately, two model variants are developed: \textbf{ChartVLM-Base-7.3B} (0.3B chart image encoder \& base decoder + 7B auxiliary decoder) and \textbf{ChartVLM-Large-8.3B} (1.3B chart image encoder \& base decoder + 7B auxiliary decoder). All the data we used during fine-tuning stage comes from ChartQA~\cite{masry2022chartqa}, PlotQA~\cite{methani2020plotqa}, Chart2Text~\cite{Kanthara2022CharttoTextAL}, and SimChart9K~\cite{xia2023structchart}. Besides, the ChartVLM is trained using 32 $\times$ NVIDIA Tesla A100.

\begin{table*}[!t]
\small
\centering	
\caption
{
Zero-shot results on both perception and cognition tasks. Comparison with state-of-the-art multi-modal language methods and chart-oriented large models on the test set of ChartX benchmark, where Desc. and Summ. denote that chart description and summarization task, respectively. The used evaluation metric for each task is introduced in Sec.~\ref{sec:eval}.
}
\vspace{-0.2cm}
\renewcommand{\arraystretch}{1.3}
\resizebox{1\textwidth}{!}{
\begin{tabular}	{l l |  c  c  c | c | c | c | c | c | c   }
\toprule	 	
 \multirow{3}{*}{Model} & \multirow{3}{*}{\makecell[l]{\#Params}} &\multicolumn{5}{c|}{ 
\textbf{Perception Tasks} }   & \multicolumn{4}{c}{ \textbf{Cognition Tasks}} \\
 & &    \multicolumn{3}{c}{Structural Extraction}& \multicolumn{1}{c}{Chart Type} & \multicolumn{1}{c|}{Chart Title} & \multicolumn{1}{c}{QA} &  \multicolumn{1}{c}{Chart Desc.} &  \multicolumn{1}{c}{Chart Summ.} &  \multicolumn{1}{c}{Chart Redraw.}  \\
     
& &  AP@Strict & AP@Slight & AP@High & EM & EM  & GPT-acc & GPT-score &  GPT-score & GPT-score \\
\midrule  
\textit{~~\textbf{Multi-modal Models} } \\
    LLaVA-1.5~\cite{liu2023improved} & 13B &
0.04 & 0.04 & 0.24 & 47.05 & 44.18 
& 17.19 & 1.48 & 1.29 & 0.75 
\\
    CogVLM~\cite{wang2023cogvlm} & 18B 
& 0.38 &  0.56 & 1.01 & 59.46 & 94.01 
& 28.30 & 2.21 & 1.48 & 1.38 
\\    
    CogAgent~\cite{hong2023cogagent} & 10B &
2.89 & 3.63 & 6.36& 61.11 & 96.27 
& 25.95 & 2.24 & 1.61 & 1.48
\\
    Monkey~\cite{li2023monkey} & 18B &
    0.00 & 0.00 & 0.00 & 66.84 & 94.44 
    & 21.61 & 1.85 & 1.62 & 1.24
\\
    QWen-VL~\cite{bai2023qwen} & 9.6B
& 4.18 & 5.86 & 8.99 & 69.53 & 94.62 
& 23.26 & 1.67 & 1.45 & 0.86 
\\    
    % InternVL~\cite{chen2023internvl} & - & 0.10 & 0.15 & 0.60 & 52.69 & 44.27 & 10.94 & 1.39 & 1.04 & 1.11\\
    SPHINX-V2~\cite{lin2023sphinx} & 13B
& 10.95 & 23.75 & 32.07 & 43.66 & 92.71 
& 31.16 & 1.53 & 1.39 &  0.96 
\\
GPT-4V~\cite{2023gpt4V}& - 
& \underline{20.91} & 26.00 & \underline{36.09} & 70.43 & \underline{95.22} & 33.04
& 3.17 & 3.12 & 2.63
\\
\midrule
\textit{~~\textbf{Chart-related Models} } \\	   
    Deplot~\cite{liu2022deplot} & 1.3B
& 8.89 & 19.04 & 24.08 & - & 89.84  
& -  & - & - & -
\\
    Matcha~\cite{liu2022matcha} & 0.3B
& 0.92 & 1.10  & 1.16 & 5.03 & 7.90  
& 14.41  & - & - & -
\\
ChartLlama~\cite{han2023chartllama} & 13B
& 1.63 & 2.01 & 3.19 & 50.52 & 40.36 
& 13.80  & 1.04 & 1.02 & 0.94
\\
% 	VILLA~\cite{villa} & & 4M & - & - & - & - & - & -  &87.9 & 97.5 & 98.8 & 76.3 & 94.2 & 96.8 \\
StructChart~\cite{xia2023structchart} & 1.3B 
& 0.46 & 0.94 & 1.77 & - & - 
& - & - & - &  - \\
ChartAst~\cite{meng2024chartassisstant} & 13B 
&11.35 & 22.77 & 30.18 & 43.23 & 92.71 
& 30.99 & 0.33 & 1.03 &  0.82 
\\	
\midrule
\multicolumn{2}{l}{\textit{~~\textbf{Ours} }} \\ 
% \textbf{ChartVLM-B} & 7.3B 
% & 18.49 & \underline{26.02} & 32.65 & \underline{95.67} & 94.27 
% & \underline{36.46} & 2.05 & 1.84 & 1.36  \\

% \textbf{ChartVLM-L} & 14.3B
%  & \textbf{23.18} & \textbf{30.68} & \textbf{38.30} & \textbf{96.82} & \textbf{97.05} 
% & \textbf{40.71} &  2.17 & \underline{2.05} & \underline{1.58} \\  

% ChartVLM-B & 8.3B 
% & 18.49 & \underline{26.02} & 32.65 & \underline{95.67} & 94.27 
% &\underline{40.01}  & \underline{3.36} &  \underline{3.32} &\underline{3.11}  \\

% ChartVLM-L & 9.3B
%  & \textbf{23.18} & \textbf{30.68} & \textbf{38.30} & \textbf{96.82} & \textbf{97.05} 
% &\textbf{42.71}  & \textbf{3.53}   & \textbf{3.43}  &\textbf{3.38}  \\ 

ChartVLM-B   & 7.3B 
& 18.49 & \underline{26.02} & 32.65 & \underline{95.67} & 94.27 
&\underline{40.19}  & \underline{3.61} &  \underline{3.43} &\underline{3.63}  \\

ChartVLM-L   & 8.3B
 & \textbf{23.18} & \textbf{30.68} & \textbf{38.30} & \textbf{96.82} & \textbf{97.05} 
&\textbf{43.84}  & \textbf{3.68}   & \textbf{3.50}  &\textbf{3.75}  \\ 

\bottomrule
\end{tabular}
}
\label{tab:compared_sota}
\end{table*}	

\begin{figure}[!t]
    \centering
    \small
    \subfloat[Structural Extraction.]{
    \begin{minipage}{0.453\linewidth}{\begin{center}
    \resizebox{\linewidth}{!}{\includegraphics{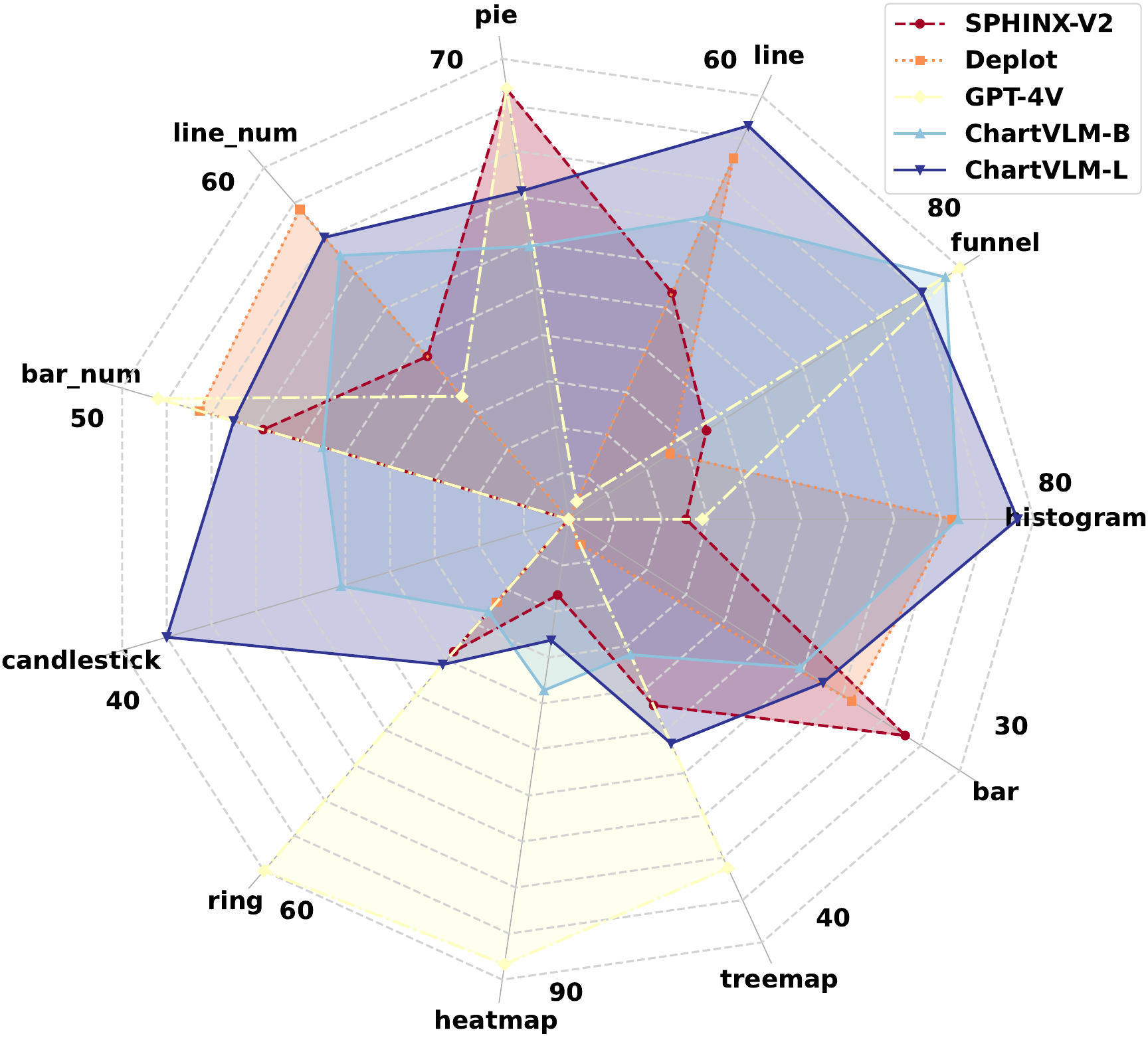}}\end{center}}\end{minipage}
    % \vspace{-10pt}
    \label{fig:exp_se}}
    \subfloat[Question-Answering.]{
    \begin{minipage}{0.45\linewidth}{\begin{center}
    \resizebox{\linewidth}{!}{\includegraphics{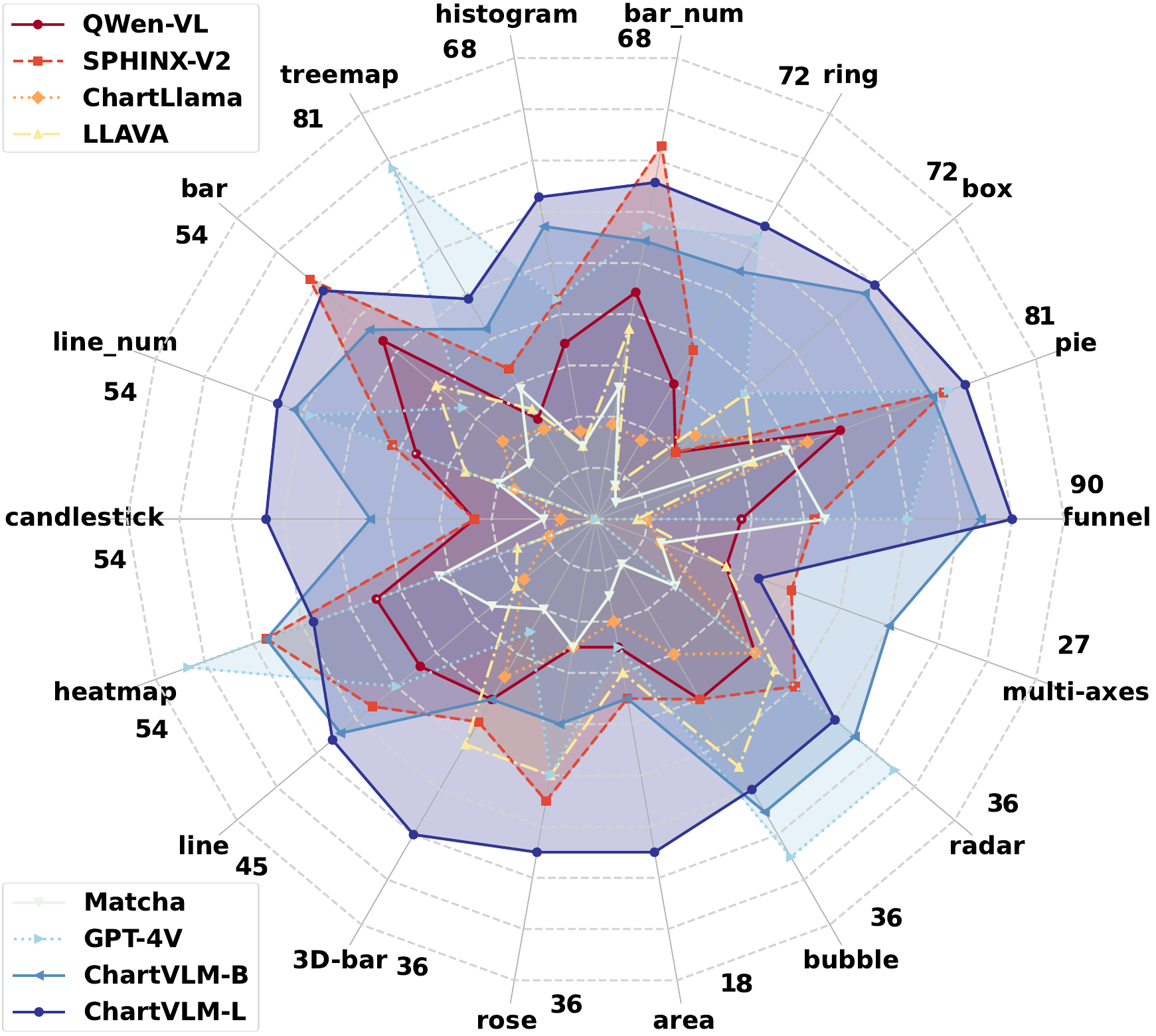}}\end{center}}\end{minipage}
    \label{fig:exp_qa}}
    \\
    \subfloat[Description.]{
    % \resizebox{\linewidth}{!}{
    \begin{minipage}{0.45\linewidth}{\begin{center}
    \resizebox{\linewidth}{!}{\includegraphics{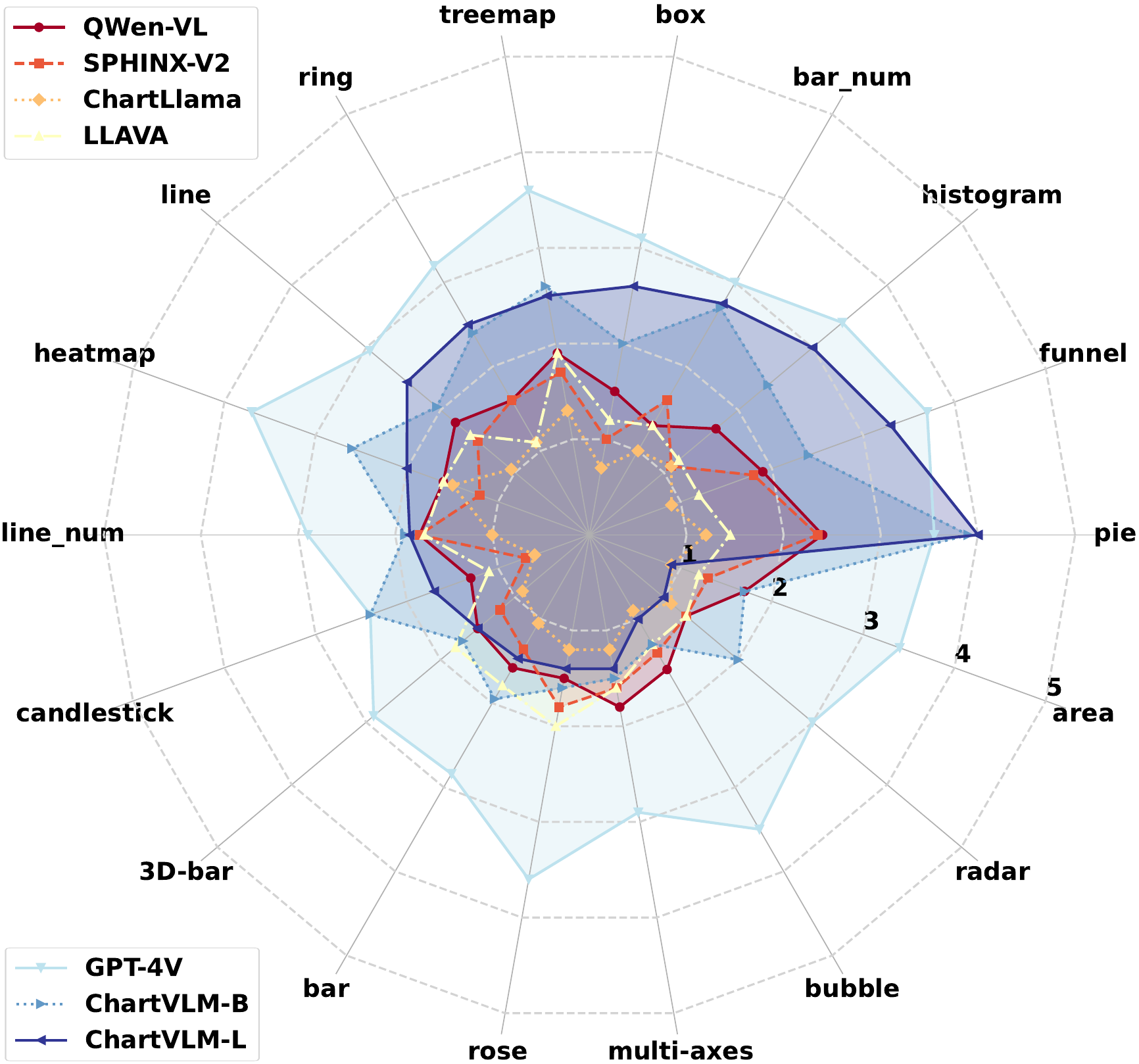}}\end{center}}\end{minipage}
    % \vspace{10pt}
    \label{fig:exp_summ}
    }
    \subfloat[Summarization.]{
    % \resizebox{\linewidth}{!}{
    \begin{minipage}{0.45\linewidth}{\begin{center}
    \resizebox{\linewidth}{!}{\includegraphics{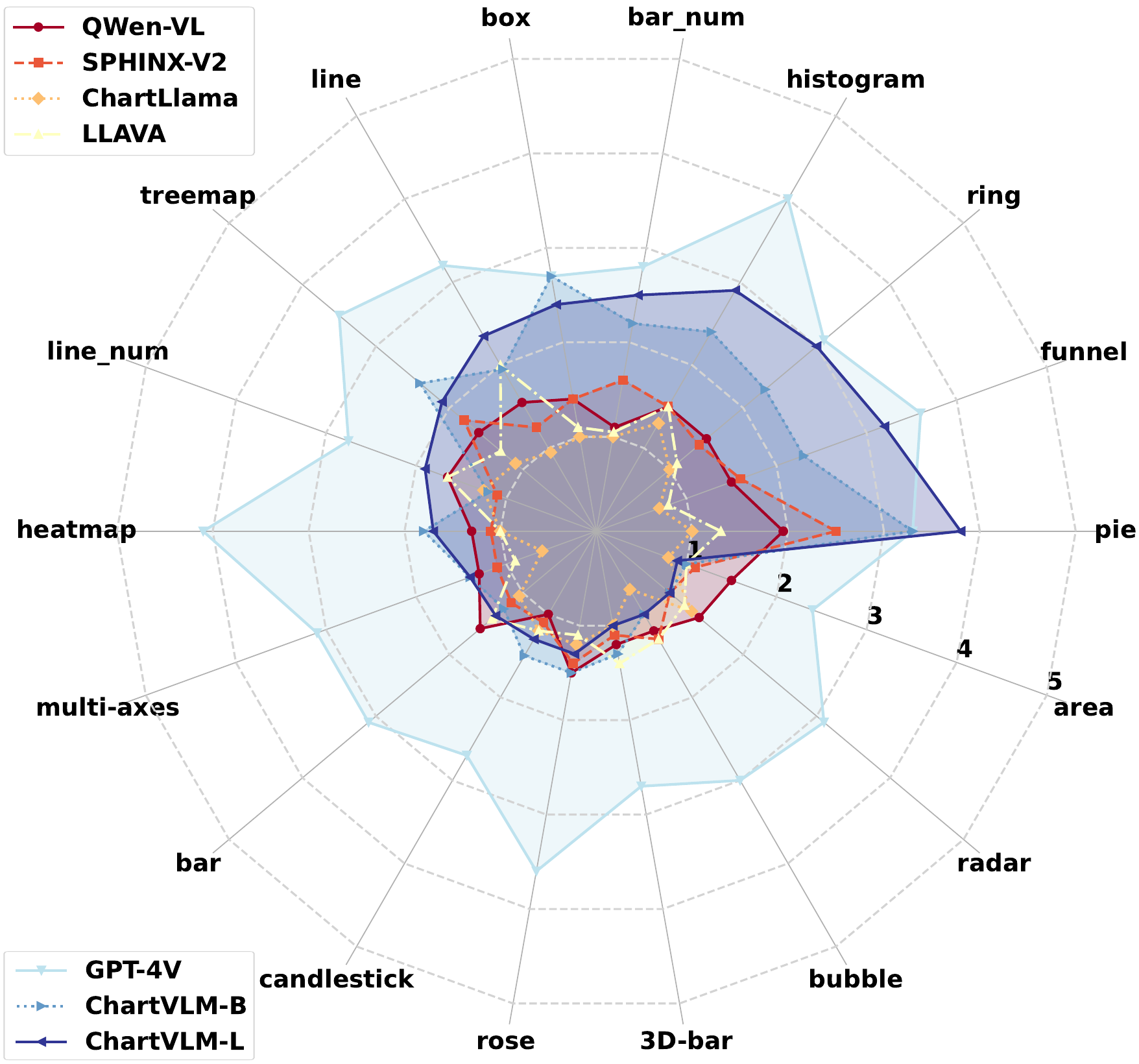}}\end{center}}\end{minipage}
    % \vspace{10pt}
    \label{fig:exp_desc}
    }
    \vspace{-2pt}
    \caption{Class-wise results of MLLMs on ChartX.}
    \label{fig:class-wise}
\vspace{-8pt}
\end{figure}

% \vspace{-2pt}
\section{Experiments}
% In this section, we first describe the experimental settings, including baseline models and the evaluation process for each task. Then, the main evaluation results of different MMLMs will be presented to demonstrate the effectiveness of our ChartX benchmark and ChartVLM model, followed by a detailed comparison between different chart types. Finally, we compare the results of our two-stage reasoning paradigm with GPT-4V and provide more insightful analyses. The implementation details of our ChartVLM are introduced in Appendix. \ref{}.

\vspace{-2pt}
\subsection{Evaluation Settings}
\vspace{-2pt}

Considering the diversity in different chart types and downstream tasks, the evaluation process of each task should be meticulously designed. Here, we first divide the ChartX benchmark into two different sets, and then describe each necessary post-processing of model predictions on different chart tasks to achieve a more objective evaluation and comparison.

\noindent\textbf{Validation Set and Test Set.} We divide the entire ChartX benchmark into 4,848 validation samples and 1,152 test samples. Both the validation and test samples underwent rigorous manual quality inspection as introduced in Appendix~\ref{para:validation}. The 4,848 validation samples are employed to evaluate the model performance during the fine-tuning process. \textbf{We report the performance of all models on the 1,152 test samples.}

\noindent\textbf{Post-processing of Structural Extraction.}
For the evaluation of the SE task, considering that the mechanism of SCRM is based on triplet-format matching and some entities may be invisible or irrelative to the visual data in some chart types, the perceived data of several chart types should be post-processed to avoid the prediction errors induced by meaningless perceptions. Specifically, for the percentage-related chart types, \textit{e.g.}, pie chart, ring chart, treemap, funnel chart, \textit{etc.}, the column label of values is usually invisible. Thus, the prediction of this entity for all task evaluations will be manually replaced as `value' or `percentage' to uniform the value representation, namely \textbf{\textit{entity replacement}}.

% The other group contains the chart types that require the special formulation of CSV data, e.g., box plot and candlestick. Specifically, for box plots, effective CSV-format data that can be explicitly recognized from the chart image and utilized for chart drawing are the lists of minimum, median, maximum, Q1, Q3, and outlier points, which usually have no explicit labels in the legend of box plots. Therefore, we uniform the labels of these six value columns as `Min', `Q1', `Median', `Q3', `Max', and `Outlier'. Similarly, for candlestick charts, the labels of four value columns are uniformly marked as `Opening Price (\$)', `Closing Price (\$)', `High Price (\$)', and `Low Price (\$)'.

\noindent\textbf{Prompt Setting for Evaluation.}
To make a fair comparison between different model performances on the ChartX benchmark, the prompts of different tasks are fine-tuned according to different baseline models to achieve the best performance on each task. The detailed prompt content for each task is illustrated in Fig.~\ref{fig:gpt_eval_1} and~\ref{fig:gpt_eval_2} of Appendix~\ref{sec:eval_setting}.

\subsection{Baseline Models and Main Results}
\vspace{-2pt}

We select two kinds of MLLMs to make a comprehensive comparison. One group of MLLMs is made up of multi-modal large models, where models are trained towards general capability for various vision-language tasks. Here we select five of the most advanced MLLMs for evaluation comparison: LLaVA-1.5 \cite{liu2023improved}, CogVLM \cite{wang2023cogvlm}, QWen-VL \cite{bai2023qwen}, SPHINX-V2 \cite{lin2023sphinx}, and GPT-4V \cite{2023gpt4V}. The other group of MLLMs represents the chart-related large models that are especially fine-tuned on chart-related tasks, including Deplot \cite{liu2022deplot}, Matcha \cite{liu2022matcha}, StructChart \cite{xia2023structchart}, ChartLlama \cite{han2023chartllama}, and ChartAssistant \cite{meng2024chartassisstant}.

Table~\ref{tab:compared_sota} shows the main comparison results with various models on test set of ChartX benchmark, from which we can observe the comprehensive evaluation results for each model across various chart tasks and the superiority of ChartVLM. Notably, the proposed ChartVLM-B and ChartVLM-L consistently outperform most models in these tasks (except GPT-4V in the cognition tasks), showcasing the effectiveness of ChartVLMs in understanding information from charts.
% Among two groups of open-source MMLMs, GPT-4V~\cite{yang2023dawn} achieves the highest performance on all seven tasks, while most of the other MMLMs significantly fall back on all tasks. The large performance gap between GPT-4V and other models demonstrates the effectiveness of ChartX in evaluating the general chart understanding ability of current chart-related large models, which will be further discussed in Sec. \ref{}. 

% \input{tables/each_class}
% \vspace{-2pt}
\textbf{Results on Each Chart Type.} The class-wise performance of ChartVLMs in seven tasks is shown in Fig.~\ref{fig:class-wise}. For better visualization, we skip six relatively difficult chart types (rose chart, area chart, 3D-bar chart, bubble chart, multi-axes chart, and radar chart) whose performance is zero-value in all AP metrics for almost all models. The numerical accuracy of these models on seven tasks can be referred to Appendix~\ref{sec:each_class}. From the four subfigures, it can be observed that the type-wise performance of different compared models and our ChartVLM can give a better understanding of different model performances across different chart types. 

% \vspace{-2pt}
\textbf{Comparison with GPT-4V.}
As shown in Table \ref{tab:compared_sota}, among all models, GPT-4V~\citep{2023gpt4V} is the only model that outperforms our ChartVLM in a few cognition tasks of the ChartX benchmark. This result is reasonable as GPT-4V is currently regarded as the most powerful MLLM for its strong ability to understand and describe information from images, \textit{e.g.}, summarization ability and description ability. However, for the perception tasks, since GPT-4V is a relatively general model, the structural extraction ability is inferior to our ChartVLM, which is specially designed for chart-related tasks. Furthermore, ChartVLM's stronger ability to extract structural data from a chart image partially leads to a higher accuracy on the chart QA task (40.71\%).

\noindent \textbf{Results using Different Auxiliary Decoder.} Based on our designed dynamic adapter mechanism, ChartVLM is capable of adapting to a variety of large language models. Here, we design an experiment where the ChartVLM employs Vicuna and Llama-3.1-8B as the fine-tuned auxiliary decoder in Table~\ref{tab:auxiliary_deocoder}. Although the performance on all cognition tasks did not surpass the version using fine-tuned Qwen2.5-7B as the auxiliary decoder, our adapted ChartVLM still managed to outperform GPT-4V in terms of accuracy on the QA tasks.

\noindent \textbf{Results on ChartQA.} We evaluate ChartVLM on the ChartQA validation set to demonstrate its generalization ability in question-answering (QA) tasks. As shown in Table~\ref{tab:chart_qa}, using Relaxed-acc as the evaluation metric, ChartVLM consistently outperforms all baseline models, including GPT-4V, similar to its performance on ChartX. Specifically, ChartVLM-B achieves an accuracy of 76.7\%, while ChartVLM-L further improves the accuracy to 77.2\%.

\begin{table*}[!t]
\small
\centering	
\caption
{
Performance of ChartVLM on cognition tasks when perform the fine-tuning using different large language models as auxiliary decoders.
}
\setlength\tabcolsep{2pt}
\resizebox{0.86\textwidth}{!}{
\begin{tabular}	{l l | c | c | c | c | c  }
\toprule	 	
 \multirow{3}{*}{Model} & \multirow{3}{*}{\makecell[l]{\#Params}} &\multirow{3}{*}{\makecell[l]{Auxiliary Decoder FT @}} & \multicolumn{4}{c}{ \textbf{Cognition Tasks}} \\
 & &  & \multicolumn{1}{c}{QA} &  \multicolumn{1}{c}{Chart Desc.} &  \multicolumn{1}{c}{Chart Summ.} &  \multicolumn{1}{c}{Chart Redraw.}  \\
     
& &   & GPT-acc & GPT-score &  GPT-score & GPT-score \\
\midrule  

GPT-4V~\cite{2023gpt4V}& - 
& - & 33.04
& 3.17 & 3.12 & 2.63
\\
\midrule

\multirow{6}{*}{ChartVLM} & 7.3B 
& Vicuna-7B
& 36.46 & 2.05 & 1.84 & 1.36  \\

 & 14.3B
 & Vicuna-13B 
& 40.71 &  2.17 & 2.05 & 1.58 \\  

 & 8.3B 
& Llama-3.1-8B
&40.01  & 3.36 &  3.32 &3.11  \\

 & 9.3B
 & Llama-3.1-8B 
& 42.71  &3.53   &3.43  & 3.38  \\ 

 & 7.3B 
& Qwen2.5-7B
&40.19  &3.61  & 3.43  &3.63  \\

 & 8.3B
 & Qwen2.5-7B
&\textbf{43.84}  &\textbf{3.68} &\textbf{3.50} &\textbf{3.75}   \\ 

\bottomrule
\end{tabular}
}
\label{tab:auxiliary_deocoder}
\end{table*}	

\begin{table}[!tb]

\centering
\footnotesize
\renewcommand{\arraystretch}{1.0}
\caption{ Question-Answering (QA) results on ChartQA val set. Relaxed-acc metric is employed for evaluation. }

\begin{tabular}{llcccccccccccccccc}
\toprule
&\multirow{2}*{Model} 
& \multicolumn{3}{c}{ChartQA val}  \\  \cmidrule{3-5}  
& & aug. \textcolor{blue!30}{\rule{0.7em}{0.55em}} & human \textcolor{red!30}{\rule{0.7em}{0.55em}} & avg.   \\

\midrule
  \multirow{6}{*}{\rotatebox[origin=c]{90}{ \textbf{Baseline} }} 
& VL-T5-OCR~\cite{masry2022chartqa}   & - & - & 41.6  \\
& Tapas-OCR~\cite{masry2022chartqa} & - & - & 45.5  \\ 
& Pix2Struct~\cite{lee2023pix2struct}  & 81.6 & 30.5 & 56.0  \\ 
& MatCha~\cite{liu2022matcha}  & 90.2  & 38.2 & 64.2  \\
& Deplot~\cite{liu2022deplot} & 69.3  & 36.6 & 52.9  \\
% &OneChart &C.D. & & & & &85.3 &49.1 &67.2 \\
&ChartLlama~\cite{han2023chartllama} & \textbf{90.4} &49.0 &69.7 \\
&GPT-4V~\cite{2023gpt4V} &76.1 &64.5 &70.3  \\
\midrule 
 \multirow{2}{*}{\rotatebox[origin=c]{90}{\textbf{Ours}}}
 % &ChartVLM-B & 82.6 & 69.5 & 76.0     \\
 % &ChartVLM-L &83.0 &\textbf{71.4} &\textbf{77.2}   \\
 & ChartVLM-B & 83.3 & 70.1 & 76.7     \\
 & ChartVLM-L & 83.5 &\textbf{70.9} &\textbf{77.2}   \\
\bottomrule 
\end{tabular}
\label{tab:chart_qa}
\end{table}

\begin{table}[!t]
% \footnotesize
\small
\renewcommand{\arraystretch}{1.1}
\centering
\caption{Accumulated prediction errors of structural extraction task towards other downstream reasoning tasks such as chart QA and chart summarization.}
\label{tab:abl_accumu}
\resizebox{0.72\linewidth}{!}{
\begin{tabular}{cccc}
    \toprule
    \multirow{2}{*}{Method} & \multirow{2}{*}{CSV Source} & QA Task & Chart Summ. \\
     & & Metric: \texttt{GPT-acc}  &  Metric: \texttt{GPT-score}  \\ 
    \hline
    ChartVLM-B & Golden Table &  52.32 & 3.62 \\
    ChartVLM-B & Predicted & 40.19  & 3.43 \\ 
    GPT-4V~\cite{2023gpt4V} & / & 33.04   & 3.12 \\
    \bottomrule
\end{tabular}}
\end{table}

\begin{table}[!t]
    % \aboverulesep = 0.48mm
    % \belowrulesep = 0.48mm
    \small
	\centering	
    \renewcommand{\arraystretch}{1.1}
    \caption{Inference speed for both perception and cognition tasks tested on a single Tesla A100 with batch size of 1. The maximum number of tokens generated for each task remains consistent.}
	\resizebox{0.80\textwidth}{!}{
	\begin{tabular}	{l | c  c  c  c |  c  c  c  c  c}
	\toprule
	\multirow{2}{*}{Model} & \multicolumn{4}{c|}{Perception Tasks} & \multicolumn{5}{c}{Cognition Tasks}\\
 
    & SE & Title & Type & Avg. & QA & Summ. & Desc. & Redraw & Avg.  \\
	\midrule  
    ~~\textbf{\textit{Inference Speed (s):}} \\
    LLaVA-1.5~\cite{liu2023improved} & 12.29 & 0.56 & 0.41 & 4.42 & 0.99 & 3.48 & 3.50 & 11.63 & 4.90\\
    % CogVLM~\cite{wang2023cogvlm} & - & 0.19 & 0.34 & 0.72 \\ 
    QWen-VL~\cite{bai2023qwen} & 4.96 & 0.93 & 1.00 & 2.30 & \textbf{0.38} & \textbf{2.98} & \textbf{2.81} & 7.43 & \textbf{3.40} \\    
    % InternVL~\cite{chen2023internvl} & - & 0.10 & 0.11 & 0.32 \\
    SPHINX-V2~\cite{lin2023sphinx} & 5.53 & 1.51 & 1.21 & 2.75 & 1.38 & 3.96 & 4.09 & 9.73 & 4.79 \\
	%GPT-4V~\cite{2023gpt4V}&  & - & 19.00 & 27.22 \\
    \midrule
    Deplot~\cite{liu2022deplot} & 3.82 & - & - & 3.82 & - & - & - & - & -\\
    % Matcha~\cite{liu2022matcha} & - & 0 & 0.16  & 0.17  \\
	ChartLlama~\cite{han2023chartllama} & 8.13 & 0.53 & 0.42 & 3.03 & 0.48 & 4.13 & 4.35 & 13.09 & 5.51\\
    ChartAst~\cite{meng2024chartassisstant} & 55.24 & 3.55 & 1.37 & 20.05 & 3.81 & 6.06 & 6.04 & 34.14 & 12.51 \\
    
    \midrule
    ChartVLM-B (ours) & \textbf{2.28} & \textbf{0.39} & \textbf{0.25} & \textbf{0.97} & 3.41 & 5.05 & 4.90 & \textbf{5.85} & 4.80\\
    ChartVLM-L (ours) & 2.87 & 0.42 & 0.29 & 1.19 & 4.38 & 6.02 & 5.98 & 7.14 & 5.88 \\
    \bottomrule
	\end{tabular}
	}
	\label{tab:speed_diff}
\vspace{-4pt}
\end{table}

\begin{table}[!t]
    % \aboverulesep = 0.48mm
    % \belowrulesep = 0.48mm
    \small
	\centering	
    \setlength\tabcolsep{4pt}
    \caption{Evaluation results of structural extraction with or without entity replacement.}
	\resizebox{0.78\textwidth}{!}{
	\begin{tabular}	{l l |  c  c  c }
	\toprule
	\multirow{2}{*}{Model} & \multirow{2}{*}{\makecell[l]{\#Params}} &    \multicolumn{3}{c}{Structural Extraction} \\
	& &  AP@Strict & AP@Slight & AP@High \\
	\midrule  
    ~~\textbf{\textit{SCRM without Entity Replacement:}} \\
    LLaVA-1.5~\cite{liu2023improved} & 13B & 0 & 0 & 0 \\
    % CogVLM~\cite{wang2023cogvlm} & - & 0.19 & 0.34 & 0.72 \\ 
    QWen-VL~\cite{bai2023qwen} & 9.6B & 1.14 & 2.40 &  4.70 \\    
    % InternVL~\cite{chen2023internvl} & - & 0.10 & 0.11 & 0.32 \\
    SPHINX-V2~\cite{lin2023sphinx} & 13B & 4.70 & 12.46 &  18.86 \\
	GPT-4V~\cite{2023gpt4V}& - & 14.35 & 19.00 & 27.22 \\
    \midrule
    Deplot~\cite{liu2022deplot} & 1.3B & 7.03 & 16.22  & 20.76 \\
    % Matcha~\cite{liu2022matcha} & - & 0 & 0.16  & 0.17  \\
	ChartLlama~\cite{han2023chartllama} & 13B & 1.39 & 1.68  & 2.37 \\
 	ChartAst~\cite{meng2024chartassisstant} & 13B & 5.99 & 14.93  & 21.19 \\
    \midrule
    % ChartVLM-B (ours) & 8.3B & 17.78 & 23.78 & 29.09 \\
    ChartVLM-L (ours) & 14.3B & 22.38 & 29.22 & 36.77 \\
	\midrule
    \textit{~~\textbf{SCRM with Entity Replacement:}} \\  
    LLaVA-1.5~\cite{liu2023improved} & 13B & 0.04 & 0.04 & 0.24 \\
    % CogVLM~\cite{wang2023cogvlm} & - & 0.38 & 0.56 & 1.01 \\ 
    QWen-VL~\cite{bai2023qwen} & 9.6B & 4.18 & 5.86 &  8.99 \\    
    % InternVL~\cite{chen2023internvl} & - & 0.10 & 0.15 & 0.60 \\
    SPHINX-V2~\cite{lin2023sphinx} & 13B & 10.95 & 23.75 &  32.07 \\
	GPT-4V~\cite{2023gpt4V}& - & 20.91 & 26.00 & 36.09 \\
    \midrule
    Deplot~\cite{liu2022deplot} & 1.3B & 8.89 & 19.04  & 24.08 \\
    % Matcha~\cite{liu2022matcha} & - & 0.92 & 1.10  & 1.16  \\
	ChartLlama~\cite{han2023chartllama} & 13B & 1.63 & 2.01  & 3.19 \\
  	ChartAst~\cite{meng2024chartassisstant} & 13B & 11.35 & 22.77  & 30.18 \\
    \midrule
    % ChartVLM-B (ours) & 8.3B & 18.49 & 25.11 & 30.26 \\
    ChartVLM-L (ours) & 14.3B & 23.18 & 30.68 & 38.30 \\
    \bottomrule
	\end{tabular}
	}
	\label{tab:eval_diff}
\end{table}

% \vspace{-2pt}
\subsection{Insightful Analyses}
% \vspace{-2pt}
In this part, we conclude five important findings as follows:
%the efficacy of the proposed ChartVLM design is principally manifested in:

% \vspace{-1em}
\textit{1) In our cascaded decoder mechanism, increased precision in structural data extraction by the base decoder is positively correlated with improved outcomes in intricate reasoning task performance.} In Table~\ref{tab:compared_sota}, it is evident that the ChartVLM-L model outperforms ChartVLM-B in SE task, also exhibiting superior performance in intricate cognition tasks, including QA, summarization, \textit{etc}. Notably, when SE accuracy attains 100\% (corresponding to `golden table' in Table~\ref{tab:abl_accumu}), our model's performance on cognition tasks peaks, indicating a direct correlation of performance between basic perception tasks and complicated cognition tasks.

% Our ChartVLM has  open-source alternatives in accomplishing intricate cognition tasks, despite exhibiting comparable capabilities in extracting structural information
% significant disparities arise in the execution of downstream cognitive tasks
% \vspace{-1pt}
\textit{2) Our ChartVLM exhibits stronger performance in complicated reasoning tasks, owing to our reasoning tasks taking the text representations obtained by the perception task as a conditional input.} Table~\ref{tab:compared_sota} demonstrates that, despite SPHINX-V2 (32.07\%) exhibiting performance close to our ChartVLM (32.65\%) in SE task, ChartVLM still demonstrates superior reasoning performance in downstream tasks such as QA tasks (36.46 \%). This improvement mainly stems from the novel design of the cascaded decoder mechanism, in which the base decoder enhances complicated reasoning tasks by incorporating the basic perceived results. 

% \vspace{-1pt}
\textit{3) Our ChartVLM demonstrates faster inference speed while maintaining a parameter count comparable to the existing open-source models.} Table~\ref{tab:speed_diff} illustrates a comparative analysis of inference speeds between ChartVLM and other open-source models. Although the inference performance on cognitive tasks is comparable across all models, a significant enhancement in speed is observed for perception tasks in ChartVLM, which is attributed to the exclusive involvement of the lightweight base decoder.

% \vspace{-3pt}
% \textit{4) The performance of certain fine-tuned chart-specific models appears to be inferior to some general MLLMs.} As illustrated in Table~\ref{tab:compared_sota}, ChartLlaMA~\cite{han2023chartllama}, which is fine-tuned on chart data from LLaVA-1.5~\cite{liu2023improved}, has inferior performance across nearly all perception and cognition tasks compared to LLaVA-1.5. Such an inferior performance may be attributed to ChartLlaMA's potential overfitting to its training data, limiting its generalizability in all-type chart tasks.

% \vspace{-1pt}
\textit{4) The post-processing implementation of entity replacement significantly alleviates assessment biases.} As shown in Table~\ref{tab:eval_diff}, entity replacement has led to enhanced performance across all baseline models in the SE task, verifying its effectiveness in refining evaluation outcomes.

% \vspace{-1pt}
\textit{5) Current MLLMs exhibit a significant deficit in their capacity to interpret type-specific charts, yielding inferior results in downstream cognitive tasks when benchmarked against GPT-4V.} As evidenced in Tables~\ref{tab:class-wise-se},~\ref{tab:class-wise-qa},~\ref{tab:class-wise-des},~\ref{tab:class-wise-summ}, and~\ref{tab:class-wise-redraw}, the existing open-source models demonstrate markedly inferior performance in both the perception and cognition tasks of specialized chart types, such as rose, area, 3D-bar, bubble, multi-axes, and radar charts. 

\subsection{Out-of-Distribution (OOD) Evaluation}

Compared with existing chart benchmarks and the training set of ChartVLM, ChartX can be considered as an OOD evaluation benchmark from two perspectives. First, ChartX contains the richest chart types (18 types) associated with the most diverse annotations and tasks (7 tasks), which is a vital element w.r.t the OOD characteristic of a chart benchmark. Second, the training set distribution of ChartVLM has almost no overlap with the evaluation set, i.e.,  ChartX. In the training set of ChartVLM, only 20\% data are synthetic, and more than 99\% data belong to the set of {\textit{line}, \textit{bar}, \textit{pie}} charts. However, ChartX contains 18 more chart types. After testing, ChartVLM shows certain performance on the other 15 chart types in addition to the three trained types. The distribution differences of various tasks between ChartX and the datasets for training ChartVLM have also been validated in Fig. \ref{fig:dis_vis} (assuming the training and test sets are iid in existing benchmarks). Therefore, the evaluation of ChartVLM and existing chart models on ChartX can be considered out of distribution.

\section{Conclusion}

In this study, to comprehensively evaluate the chart-related capabilities of MLLMs, we construct ChartX, which is a high-quality, multi-modal, multi-type, multi-topic, and multi-task chart evaluation set. Besides, the ChartVLM framework is developed, which leverages a new cascaded decoder mechanism to boost the interpretability of MLLMs in handling scientific chart data.

% \newpage
% \section*{Impact Statements}
% The proposed ChartX is useful in many fields for chart understanding, such as medical, education, finance, business, \textit{etc}. One potential negative societal impact is: our approach may perform chart image plagiarism using the proposed chart redrawing task, which may raise privacy concerns and result in the academic misconduct. Nevertheless, our ChartVLM model can perform the Structural Extraction (SE) task well, and extract the structural representations from a chart image. On the positive side, this SE task can be used for chart duplication checking.

\section*{Acknowledgement}

The research was supported by Shanghai Artificial Intelligence Laboratory, the National Key R\&D Program of China (Grant No. 2022ZD0160104), the Science and Technology Commission of Shanghai Municipality (Grant No. 22DZ1100102), and Shanghai Rising Star Program (Grant No. 23QD1401000).

% The work was supported in part by Science and Technology Commission of Shanghai Municipality under Grant No. 22DZ1100102, in part by National Key R\&D Program of China under Grant No. 2022ZD0160104 and Shanghai Rising Star Program under Grant No. 23QD1401000, in part by National Natural Science Foundation of China under Grant Nos. 62222607, 61972250, and U19B2035.

{\small
% \normalem
\bibliographystyle{plainnat}
\bibliography{ref}
}

% APPENDIX
%%%%%%%%%%%%%%%%%%%%%%%%%%%%%%%%%%%%%%%%%%%%%%%%%%%%%%%%%%%%%%%%%%%%%%%%%%%%%%%
%%%%%%%%%%%%%%%%%%%%%%%%%%%%%%%%%%%%%%%%%%%%%%%%%%%%%%%%%%%%%%%%%%%%%%%%%%%%%%%
\newpage
\setcounter{equation}{0}
\setcounter{figure}{0}
\setcounter{table}{0}
\appendix
\renewcommand\thefigure{A.\arabic{figure}}
\renewcommand\theequation{A.\arabic{equation}}
\renewcommand\thetable{A.\arabic{table}}

\section{Details of ChartX Evaluation Set}
We present the zoom-in characteristics of the ChartX evaluation set by detailing the data distribution and its generation pipeline.

\subsection{Introduction of Chart Topics}
\label{sec:chart_topic}
The categories of chart topics have been concisely displayed in Fig.~\ref{fig1:motivation} of the main text. Here a more detailed distribution is introduced for a clear visualization. As shown in Fig. \ref{fig:topic}, there are a total of 22 chart topics, generally covering the fields of commerce, industry, lifestyle, society, and culture. Each topic is evenly distributed in ChartX, demonstrating its comprehensiveness.

\begin{figure}[h]
    \centering
    \resizebox{0.78\linewidth}{!}{\includegraphics{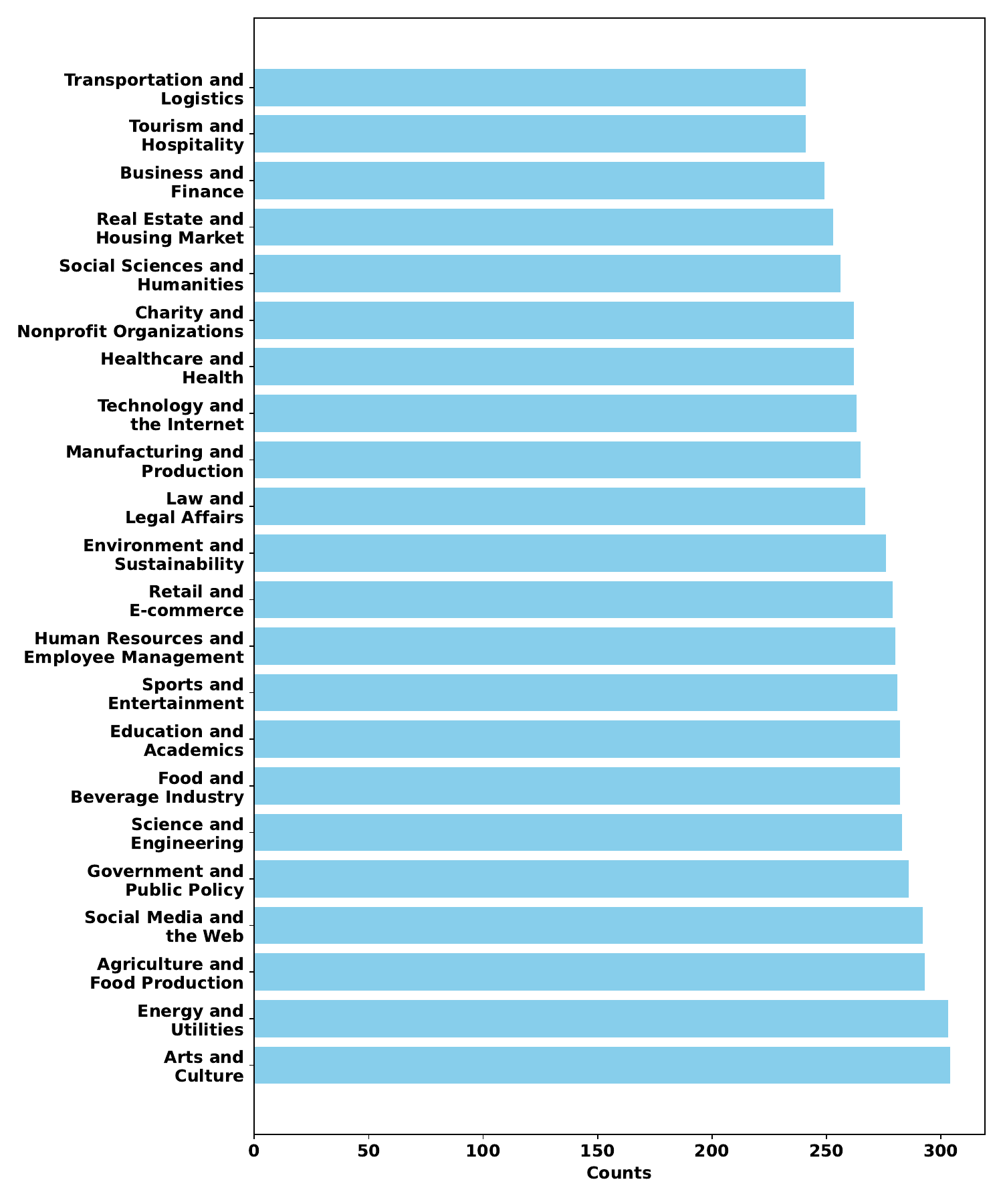}}
    % \vspace{-2em}
    \caption{The distribution of fine-grained chart topics.} 
    \label{fig:topic}
\end{figure}

\subsection{Overall of Data Generation Pipeline}
\label{sec:data_gen}
We first describe the overall data generation pipeline, including perception data and cognition data. Then, the prompt templates for different data generation are provided.

\noindent \textbf{Data Generation Pipeline.} 
As shown in Fig. \ref{fig:generation}, during the first stage, we prepare a chart type pool and a chart topic pool in which the candidates are pre-selected based on GPT-4, where those chart types of an explicit connection or mapping with CSV-format data are selected as candidates of the chart type pool. After achieving such two pools, we iteratively and randomly sample the candidates from two pools and fill them into the pre-designed prompt template to generate CSV data associated with the chart title. Once the pair of CSV data and the corresponding chart title are generated, they are both filled into various task-specific and type-specific prompt templates to generate cognition task samples.

\noindent \textbf{Prompt Design for Overall Data Generation. }
\label{sec:specific_type_gen}
We provide a general prompt template for overall data generation, including perception data and cognition data in Fig.~\ref{fig:gen_tem_1}. For perception data generation, we impose constraints on the magnitude and length of the data to make most data visible and recognizable in the chart image. For cognition data generation, we impose task-specific guidance to generate the corresponding ground-truth labels for each task. The diversity in different tasks is achieved through designing type-specific prompts. Here we provide two examples to illustrate type-specific prompts (marked red in Fig.~\ref{fig:gen_tem_1}) in overall data generation. Fig.~\ref{fig:gen_tem_2} shows the detailed type-specific prompts to generate code data and QA samples of 3D-bar charts, rose charts, box plots and candlesticks.

\vspace{-8pt}
\begin{figure*}[!t]
    \centering
    \resizebox{0.93\linewidth}{!}{\includegraphics{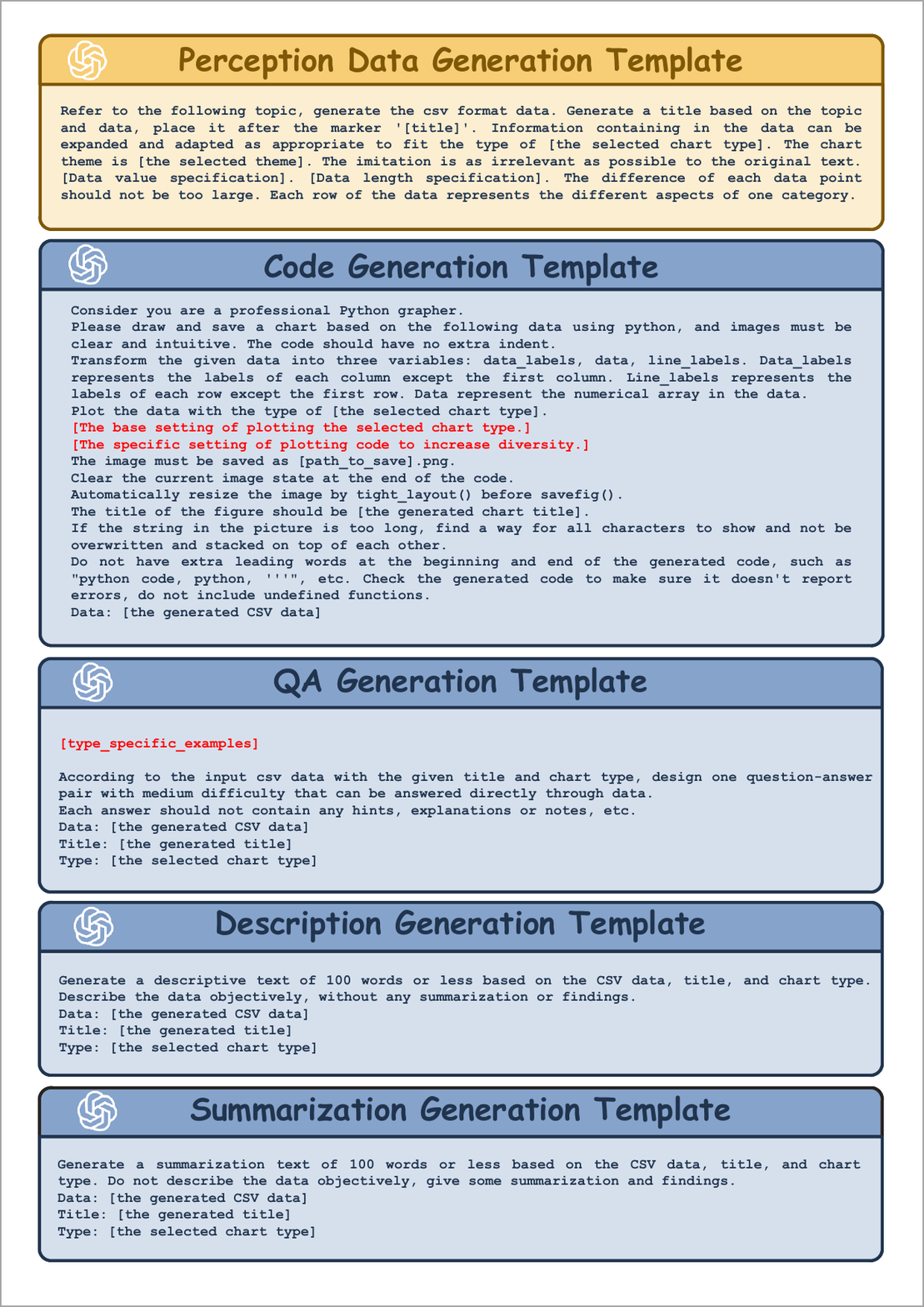}}
    % \vspace{-8pt}
    \caption{Prompts designed for overall data generation, including perception data and cognition data. The content marked red refers to the type-specific prompt, which will be illustrated in Fig.~\ref{fig:gen_tem_2}.} 
    \label{fig:gen_tem_1}
    \vspace{-4pt}
\end{figure*}

\vspace{-2pt}
\begin{figure*}[!t]
    \centering
    \resizebox{0.95\linewidth}{!}{\includegraphics{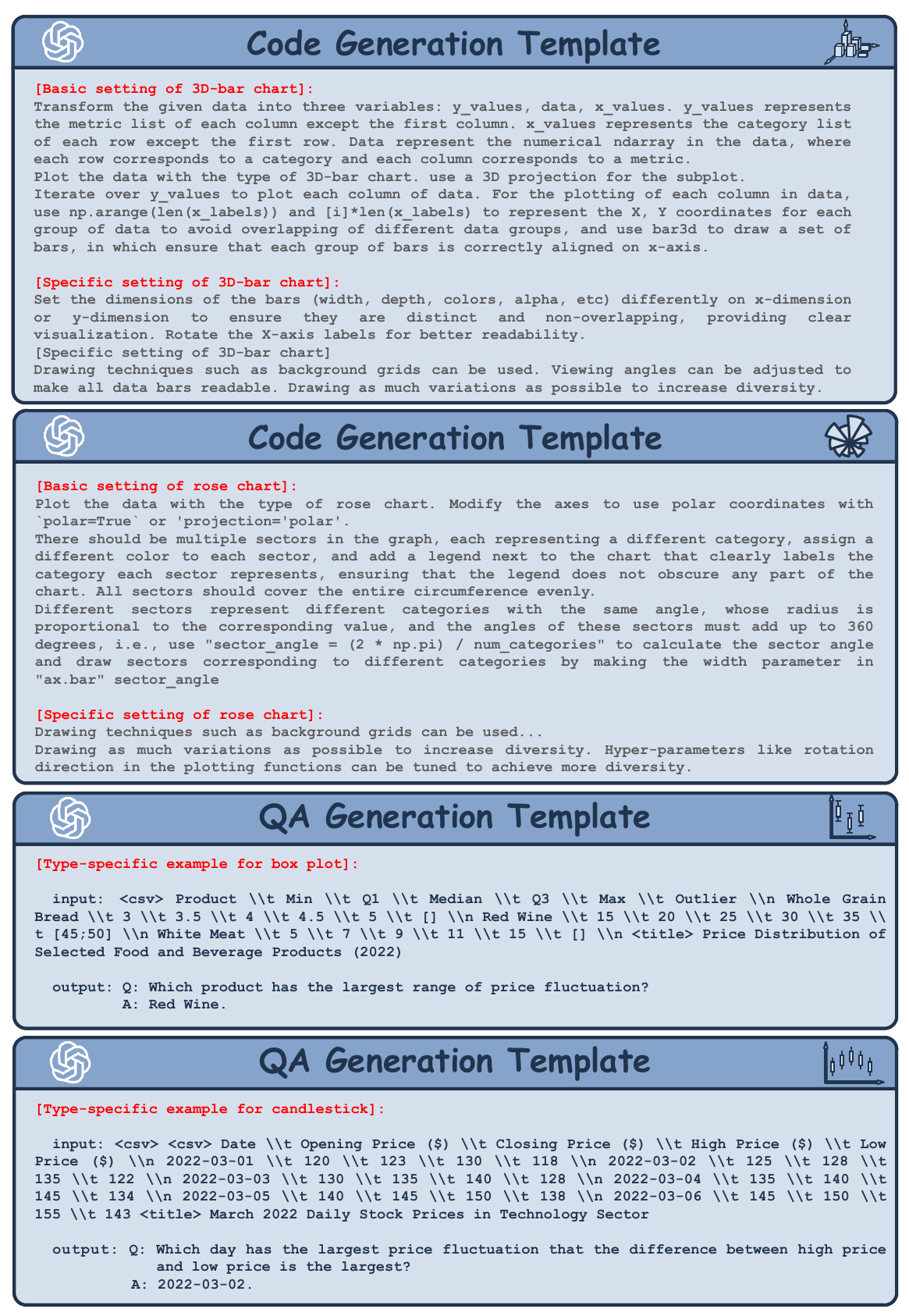}}
    % \vspace{-8pt}
    \caption{Examples of type-specific prompt design for 3D-bar chart, rose chart, box plot and candlesticks.} 
    \label{fig:gen_tem_2}
    \vspace{-2pt}
\end{figure*}

\subsection{Manual Quality Inspection}
\label{para:validation}
The human validation process is implemented by a team group professional in chart processing. The validation process for each data sample in ChartX went through four steps.

\textbf{Step One.} Overall Format Validation by Programming. After the generation of chart image, CSV-format data, text-format information about chart title and type, and the corresponding drawing codes. We first cleared the pairs that contain null files by python programming to check whether there exist null files that failed in saving the generated data. The null ratio is under 1\% in this validation step. Then, we checked the correctness of the saving format in each modality, especially the CSV format. This is achieved by calling the ‘read\_csv’ function in Pandas package and testing whether the data length is non-positive. The detection rate of failure cases in this sub-step is under 15\%. To this step, we ensure the saved files are format-correct.

\textbf{Step Two.} Image-CSV-txt-code Pair Content Validation by Human. Next, a team of experts meticulously reviewed the alignment and accuracy of the content across the different modalities (image, CSV, text, and code) to ensure they match and accurately represent the same data. This involves comparing the chart image against the CSV data to verify that the chart accurately visualizes the data. The text and code were also checked to ensure they correctly describe and generate the chart, respectively. The detection rate varies in different chart types due to different plotting methods. The average detection rate in this step is under 5\%.

\textbf{Step Three.} CSV-Question-Answering Pair Content Validation by Human and GPT-4. In step three, the focus shifted to the validation of question-answering pairs related to the chart. This first involves the step-by-step reasoning generated by GPT-4 according to the chart data and the given question. Then the human experts meticulously validated the rationality and correctness of each reasoning step to ensure the correctness of the generated answer and the relevance to the chart data. The detection rate in this step is under 1\%.

\textbf{Step Four.} Image-Summarization-Description Pair Content Validation by Human and GPT-4. The fourth and final step is dedicated to validating the image-summarization-description pairs. This first involves the GPT-4’s self-evaluation of both summarization and description labels based on the designed GPT-score metrics. Those with GPT-score lower than 4 will be modified by human experts according to CSV data and the chart image if there exist minor misinterpretations, otherwise the text will be regenerated by GPT-4 with human feedback until the score is higher than 4. The detection rate in this step is under 1\%.

\subsection{Examples of ChartX}

Fig.~\ref{fig:meta_data} provides more examples of metadata in the generated dataset, including the chart type, title, topic, CSV data, QA pairs, summarization, description, and the redrawing code. It can be observed that:

(1) The generated data are closely related to the assigned chart types and topics.

(2) The generated QA pairs are closely related to the characteristics of the given chart types and topics, increasing the overall diversity.

(3) The generated summary and description concisely and accurately describe the content of the assigned chart data.

\vspace{-2pt}
\begin{figure*}[!t]
    \centering
    \resizebox{0.95\linewidth}{!}{\includegraphics{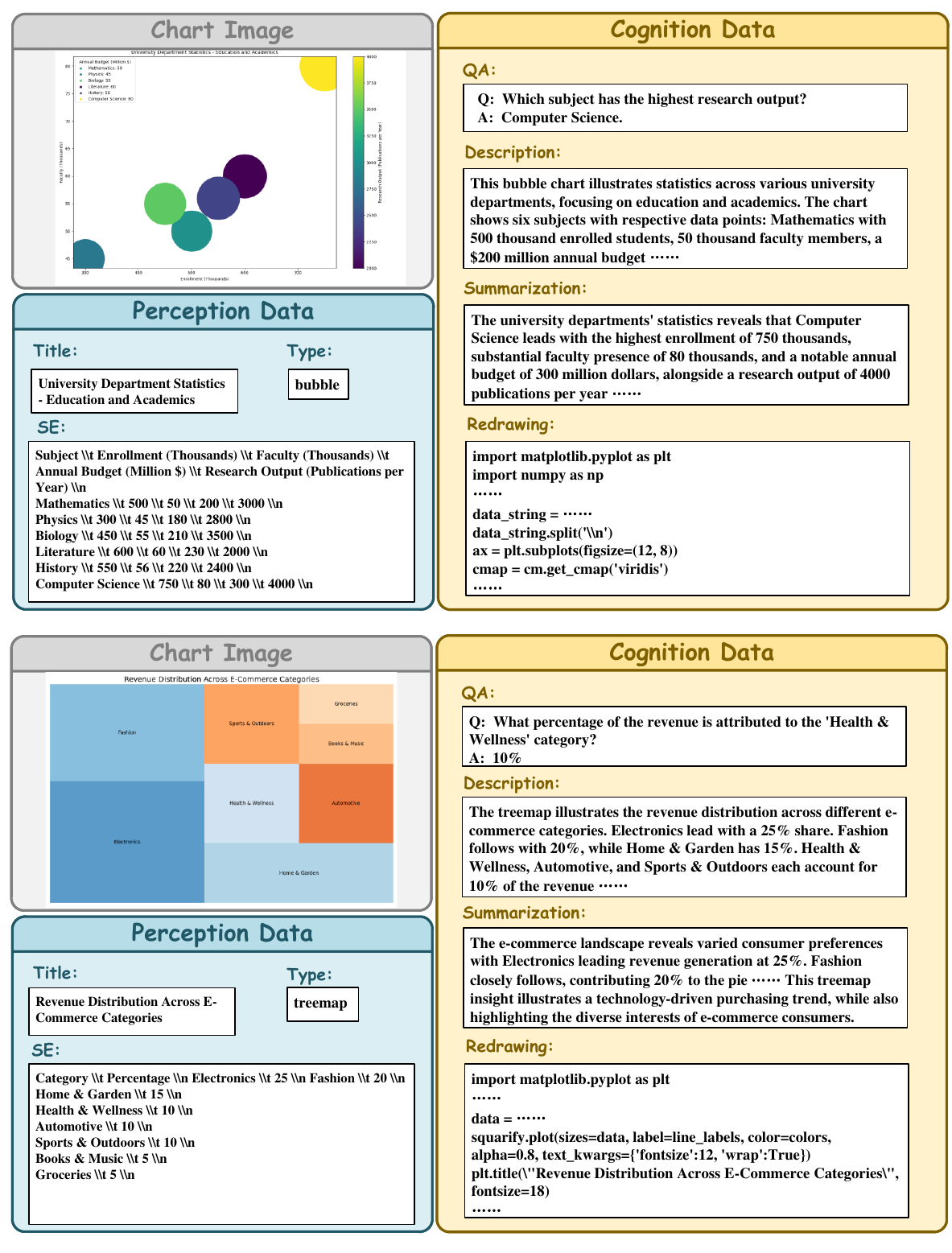}}
    % \vspace{-8pt}
    \caption{Two examples of metadata in ChartX, including the chart type, title, topic, CSV data, QA pairs, summarization, description, and the redrawing code.} 
    \label{fig:meta_data}
    \vspace{-8pt}
\end{figure*}

\section{Experimental Details}
We provide detailed experimental information in this section, including the evaluation criteria of all tasks, the quantitative results for each chart type, and more visualizations of prediction results.

\subsection{Training Configuration}
\textbf{For the chart extraction model}, the training set of four datasets are used for fine-tuning: ChartQA, Chart2text, PlotQA, and SimChart9K, where the CSV data are labeled as ground-truth for the finetuning of the encoder and base decoder.
\textbf{For the LLM decoder}, the training set of four datasets are used for fine-tuning: ChartQA, Chart2text, PlotQA, and SimChart9K, where all available QA pair, summarization data and python codes are selected as the instruction-following dataset to fine-tune the LLM decoder with the alpaca format. QA data from ChartQA and PlotQA, summarization data from Chart2text and SimChart9K, and redrawing code data from SimChart9K are employed for fine-tuning the LLM decoder.
\textbf{For the instruction Adapter}, we generate 7K pairs of user instructions and the corresponding task category based on GPT (version: GPT-3.5-turbo).

\begin{table*}[t]\scriptsize
\renewcommand{\arraystretch}{1.6}
\centering

\caption{\textbf{Class-wise mean precision for Structural Extraction (SE) task evaluated using SCRM~\cite{xia2023structchart}}. For some hard fine-grained classes such as bubble chart, radar chart, \textit{etc}, we use the relatively high tolerance for evaluating the SCRM results as introduced in Sec.~\ref{sec:scrm_tol}. Note that the color blocks represent the tolerance level we set in SCRM, where \textcolor{red!60}{\rule{0.5em}{0.5em}}, \textcolor{yellow!60}{\rule{0.5em}{0.5em}}, \textcolor{green!60}{\rule{0.5em}{0.5em}} indicate \textit{strict, slight, high} tolerance, respectively.
}

\resizebox{\textwidth}{!}{%
\begin{tabular}{l|c|ccccc|ccc|ccc|cccc|ccc|c}
    \toprule
    &  & \multicolumn{5}{c|}{\textbf{General Chart Types}} & \multicolumn{13}{c|}{\textbf{Fine-grained Chart Types}} & \\
    \multirow{-2}{*}{Models} & \multirow{-2}{*}{Tasks} & bar & bar\_num & line & line\_num & pie &  ring & box & hist & treemap  &  rose  & area & 3D-bar &  bubble & multi & radar & heatmap & funnel & candle & \multirow{-2}{*}{Avg.} \\
    \hline
     
     % FILIP~\citep{yao2021filip} & $\times$ &
     % 89.8 & 99.2 & 99.8 & 75.0 & 93.4 & 96.3 & 61.3 & 84.3 & 90.4 & 45.9 & 70.6 & 79.3 & 82.1 \\
     \multirow{3}*{\textbf{SPHINX-V2}} & \multirow{18}*{SE} &
     \textcolor{red!60}{\rule{0.5em}{0.5em}}2.50  & \textcolor{red!60}{\rule{0.5em}{0.5em}}20.10
     & \textcolor{red!60}{\rule{0.5em}{0.5em}}7.20 &\textcolor{red!60}{\rule{0.5em}{0.5em}}9.90
     & \textcolor{red!60}{\rule{0.5em}{0.5em}}35.10 &\textcolor{red!60}{\rule{0.5em}{0.5em}}9.00
     &\textcolor{red!60}{\rule{0.5em}{0.5em}}0.00 &\textcolor{red!60}{\rule{0.5em}{0.5em}}2.00
     &\textcolor{red!60}{\rule{0.5em}{0.5em}}17.60 &\textcolor{red!60}{\rule{0.5em}{0.5em}}15.40
     &\textcolor{red!60}{\rule{0.5em}{0.5em}}0.00 &\textcolor{red!60}{\rule{0.5em}{0.5em}} 0.00
     &\textcolor{red!60}{\rule{0.5em}{0.5em}}0.00 &\textcolor{red!60}{\rule{0.5em}{0.5em}} 0.00
     &\textcolor{red!60}{\rule{0.5em}{0.5em}}0.00 &\textcolor{red!60}{\rule{0.5em}{0.5em}} 9.81
     &\textcolor{red!60}{\rule{0.5em}{0.5em}}26.00 &\textcolor{red!60}{\rule{0.5em}{0.5em}}0.00
     &\textcolor{red!60}{\rule{0.5em}{0.5em}}10.95\\
      
    &&\textcolor{yellow!60}{\rule{0.5em}{0.5em}}17.40 
    &\textcolor{yellow!60}{\rule{0.5em}{0.5em}}34.20 	
    &\textcolor{yellow!60}{\rule{0.5em}{0.5em}}36.40 	
    &\textcolor{yellow!60}{\rule{0.5em}{0.5em}}27.80 	
    &\textcolor{yellow!60}{\rule{0.5em}{0.5em}}65.40 	
    &\textcolor{yellow!60}{\rule{0.5em}{0.5em}}22.60 	
    &\textcolor{yellow!60}{\rule{0.5em}{0.5em}}0.00 	
    &\textcolor{yellow!60}{\rule{0.5em}{0.5em}}20.20 	
    &\textcolor{yellow!60}{\rule{0.5em}{0.5em}}17.60 	
    &\textcolor{yellow!60}{\rule{0.5em}{0.5em}}51.00 	
    &\textcolor{yellow!60}{\rule{0.5em}{0.5em}}0.00 	
    &\textcolor{yellow!60}{\rule{0.5em}{0.5em}}2.60 	
    &\textcolor{yellow!60}{\rule{0.5em}{0.5em}}0.00 	
    &\textcolor{yellow!60}{\rule{0.5em}{0.5em}}0.00 	
    &\textcolor{yellow!60}{\rule{0.5em}{0.5em}}8.00 	
    &\textcolor{yellow!60}{\rule{0.5em}{0.5em}}14.81 	
    &\textcolor{yellow!60}{\rule{0.5em}{0.5em}}28.20 	
    &\textcolor{yellow!60}{\rule{0.5em}{0.5em}}0.00 	
    &\textcolor{yellow!60}{\rule{0.5em}{0.5em}}23.75 \\

    &&\textcolor{green!60}{\rule{0.5em}{0.5em}}39.40 	
    &\textcolor{green!60}{\rule{0.5em}{0.5em}}46.00 	
    &\textcolor{green!60}{\rule{0.5em}{0.5em}}47.90 	
    &\textcolor{green!60}{\rule{0.5em}{0.5em}}39.70 	
    &\textcolor{green!60}{\rule{0.5em}{0.5em}}76.20 	
    &\textcolor{green!60}{\rule{0.5em}{0.5em}}27.80 	
    &\textcolor{green!60}{\rule{0.5em}{0.5em}}0.80 	
    &\textcolor{green!60}{\rule{0.5em}{0.5em}}25.60 	
    &\textcolor{green!60}{\rule{0.5em}{0.5em}}18.40 	
    &\textcolor{green!60}{\rule{0.5em}{0.5em}}71.40 	
    &\textcolor{green!60}{\rule{0.5em}{0.5em}}1.40 	
    &\textcolor{green!60}{\rule{0.5em}{0.5em}}4.60 	
    &\textcolor{green!60}{\rule{0.5em}{0.5em}}0.00 	
    &\textcolor{green!60}{\rule{0.5em}{0.5em}}0.60 	
    &\textcolor{green!60}{\rule{0.5em}{0.5em}}16.00 	
    &\textcolor{green!60}{\rule{0.5em}{0.5em}}18.46 	
    &\textcolor{green!60}{\rule{0.5em}{0.5em}}34.20 	
    &\textcolor{green!60}{\rule{0.5em}{0.5em}}0.00 	
    &\textcolor{green!60}{\rule{0.5em}{0.5em}}32.07  \\
     \cmidrule(l){3-21}
     
    \multirow{3}*{\textbf{Deplot}} &  &\textcolor{red!60}{\rule{0.5em}{0.5em}}2.20 	
    &\textcolor{red!60}{\rule{0.5em}{0.5em}}33.70 	
    &\textcolor{red!60}{\rule{0.5em}{0.5em}}16.00 	
    &\textcolor{red!60}{\rule{0.5em}{0.5em}}22.30 	
    &\textcolor{red!60}{\rule{0.5em}{0.5em}}0.00 	
    &\textcolor{red!60}{\rule{0.5em}{0.5em}}14.20 	
    &\textcolor{red!60}{\rule{0.5em}{0.5em}}0.00 	
    &\textcolor{red!60}{\rule{0.5em}{0.5em}}20.20 	
    &\textcolor{red!60}{\rule{0.5em}{0.5em}}2.40 	
    &\textcolor{red!60}{\rule{0.5em}{0.5em}}0.00 	
    &\textcolor{red!60}{\rule{0.5em}{0.5em}}0.00 	
    &\textcolor{red!60}{\rule{0.5em}{0.5em}}0.00 	
    &\textcolor{red!60}{\rule{0.5em}{0.5em}}0.00 	
    &\textcolor{red!60}{\rule{0.5em}{0.5em}}0.00 	
    &\textcolor{red!60}{\rule{0.5em}{0.5em}}0.00 	
    &\textcolor{red!60}{\rule{0.5em}{0.5em}}0.00 	
    &\textcolor{red!60}{\rule{0.5em}{0.5em}}19.60 	
    &\textcolor{red!60}{\rule{0.5em}{0.5em}}0.00 	
    &\textcolor{red!60}{\rule{0.5em}{0.5em}}8.89  \\
      
    && \textcolor{yellow!60}{\rule{0.5em}{0.5em}}21.70 	
    & \textcolor{yellow!60}{\rule{0.5em}{0.5em}}41.30 	
    & \textcolor{yellow!60}{\rule{0.5em}{0.5em}}51.20 	
    & \textcolor{yellow!60}{\rule{0.5em}{0.5em}}52.90 	
    & \textcolor{yellow!60}{\rule{0.5em}{0.5em}}0.00 	
    & \textcolor{yellow!60}{\rule{0.5em}{0.5em}}14.20 	
    & \textcolor{yellow!60}{\rule{0.5em}{0.5em}}0.00 	
    & \textcolor{yellow!60}{\rule{0.5em}{0.5em}}66.00 	
    & \textcolor{yellow!60}{\rule{0.5em}{0.5em}}2.40 	
    & \textcolor{yellow!60}{\rule{0.5em}{0.5em}}0.20 	
    & \textcolor{yellow!60}{\rule{0.5em}{0.5em}}0.20 	
    & \textcolor{yellow!60}{\rule{0.5em}{0.5em}}0.00 	
    & \textcolor{yellow!60}{\rule{0.5em}{0.5em}}0.00 	
    & \textcolor{yellow!60}{\rule{0.5em}{0.5em}}0.20 	
    & \textcolor{yellow!60}{\rule{0.5em}{0.5em}}0.40 	
    & \textcolor{yellow!60}{\rule{0.5em}{0.5em}}0.00 	
    & \textcolor{yellow!60}{\rule{0.5em}{0.5em}}20.80 	
    & \textcolor{yellow!60}{\rule{0.5em}{0.5em}}0.00 	
    & \textcolor{yellow!60}{\rule{0.5em}{0.5em}}19.04  \\

    && \textcolor{green!60}{\rule{0.5em}{0.5em}}42.10 	
    & \textcolor{green!60}{\rule{0.5em}{0.5em}}48.70 	
    & \textcolor{green!60}{\rule{0.5em}{0.5em}}60.10 	
    & \textcolor{green!60}{\rule{0.5em}{0.5em}}61.20 	
    & \textcolor{green!60}{\rule{0.5em}{0.5em}}0.00 	
    & \textcolor{green!60}{\rule{0.5em}{0.5em}}14.60 	
    & \textcolor{green!60}{\rule{0.5em}{0.5em}}0.00 	
    & \textcolor{green!60}{\rule{0.5em}{0.5em}}82.20 	
    & \textcolor{green!60}{\rule{0.5em}{0.5em}}3.00 	
    & \textcolor{green!60}{\rule{0.5em}{0.5em}}4.20 	
    & \textcolor{green!60}{\rule{0.5em}{0.5em}}0.60 	
    & \textcolor{green!60}{\rule{0.5em}{0.5em}}0.00 	
    & \textcolor{green!60}{\rule{0.5em}{0.5em}}0.00 	
    & \textcolor{green!60}{\rule{0.5em}{0.5em}}1.40 	
    & \textcolor{green!60}{\rule{0.5em}{0.5em}}1.00 	
    & \textcolor{green!60}{\rule{0.5em}{0.5em}}0.00 	
    & \textcolor{green!60}{\rule{0.5em}{0.5em}}23.60 	
    & \textcolor{green!60}{\rule{0.5em}{0.5em}}0.00 	
    & \textcolor{green!60}{\rule{0.5em}{0.5em}}24.08  \\
    \cmidrule(l){3-21}

    \multirow{3}*{\textbf{ChartAst}} &  &\textcolor{red!60}{\rule{0.5em}{0.5em}}7.80 	
    &\textcolor{red!60}{\rule{0.5em}{0.5em}}22.10 	
    &\textcolor{red!60}{\rule{0.5em}{0.5em}}8.20 	
    &\textcolor{red!60}{\rule{0.5em}{0.5em}}11.50 	
    &\textcolor{red!60}{\rule{0.5em}{0.5em}}44.30 	
    &\textcolor{red!60}{\rule{0.5em}{0.5em}}4.40 	
    &\textcolor{red!60}{\rule{0.5em}{0.5em}}0.00 	
    &\textcolor{red!60}{\rule{0.5em}{0.5em}}8.40 	
    &\textcolor{red!60}{\rule{0.5em}{0.5em}}13.60 	
    &\textcolor{red!60}{\rule{0.5em}{0.5em}}2.40 	
    &\textcolor{red!60}{\rule{0.5em}{0.5em}}0.00 	
    &\textcolor{red!60}{\rule{0.5em}{0.5em}}0.00 	
    &\textcolor{red!60}{\rule{0.5em}{0.5em}}1.40 	
    &\textcolor{red!60}{\rule{0.5em}{0.5em}}0.00 	
    &\textcolor{red!60}{\rule{0.5em}{0.5em}}0.00 	
    &\textcolor{red!60}{\rule{0.5em}{0.5em}}13.65 	
    &\textcolor{red!60}{\rule{0.5em}{0.5em}}9.20 	
    &\textcolor{red!60}{\rule{0.5em}{0.5em}}0.00 	
    &\textcolor{red!60}{\rule{0.5em}{0.5em}}11.35  \\
      
    && \textcolor{yellow!60}{\rule{0.5em}{0.5em}}21.70 	
    & \textcolor{yellow!60}{\rule{0.5em}{0.5em}}33.80 	
    & \textcolor{yellow!60}{\rule{0.5em}{0.5em}}40.10 	
    & \textcolor{yellow!60}{\rule{0.5em}{0.5em}}35.20 	
    & \textcolor{yellow!60}{\rule{0.5em}{0.5em}}53.00 	
    & \textcolor{yellow!60}{\rule{0.5em}{0.5em}}14.80 	
    & \textcolor{yellow!60}{\rule{0.5em}{0.5em}}0.00 	
    & \textcolor{yellow!60}{\rule{0.5em}{0.5em}}24.80 	
    & \textcolor{yellow!60}{\rule{0.5em}{0.5em}}14.60 	
    & \textcolor{yellow!60}{\rule{0.5em}{0.5em}}25.20 	
    & \textcolor{yellow!60}{\rule{0.5em}{0.5em}}0.00 	
    & \textcolor{yellow!60}{\rule{0.5em}{0.5em}}3.80 	
    & \textcolor{yellow!60}{\rule{0.5em}{0.5em}}1.80 	
    & \textcolor{yellow!60}{\rule{0.5em}{0.5em}}0.00 	
    & \textcolor{yellow!60}{\rule{0.5em}{0.5em}}26.00 	
    & \textcolor{yellow!60}{\rule{0.5em}{0.5em}}20.00 	
    & \textcolor{yellow!60}{\rule{0.5em}{0.5em}}11.20 	
    & \textcolor{yellow!60}{\rule{0.5em}{0.5em}}0.00 	
    & \textcolor{yellow!60}{\rule{0.5em}{0.5em}}22.77  \\

    &&\textcolor{green!60}{\rule{0.5em}{0.5em}}38.40 	
    &\textcolor{green!60}{\rule{0.5em}{0.5em}}44.60 	
    &\textcolor{green!60}{\rule{0.5em}{0.5em}}48.00 	
    &\textcolor{green!60}{\rule{0.5em}{0.5em}}41.70 	
    &\textcolor{green!60}{\rule{0.5em}{0.5em}}63.70 	
    &\textcolor{green!60}{\rule{0.5em}{0.5em}}14.80 	
    &\textcolor{green!60}{\rule{0.5em}{0.5em}}0.00 	
    &\textcolor{green!60}{\rule{0.5em}{0.5em}}30.80 	
    &\textcolor{green!60}{\rule{0.5em}{0.5em}}15.80 	
    &\textcolor{green!60}{\rule{0.5em}{0.5em}}40.60 	
    &\textcolor{green!60}{\rule{0.5em}{0.5em}}0.00 	
    &\textcolor{green!60}{\rule{0.5em}{0.5em}}7.00 	
    &\textcolor{green!60}{\rule{0.5em}{0.5em}}4.20 	
    &\textcolor{green!60}{\rule{0.5em}{0.5em}}0.00 	
    &\textcolor{green!60}{\rule{0.5em}{0.5em}}38.00 	
    &\textcolor{green!60}{\rule{0.5em}{0.5em}}24.04 	
    &\textcolor{green!60}{\rule{0.5em}{0.5em}}15.00 	
    &\textcolor{green!60}{\rule{0.5em}{0.5em}}0.00 	
    &\textcolor{green!60}{\rule{0.5em}{0.5em}}30.18  \\
    \cmidrule(l){3-21}

    \multirow{3}*{\textbf{GPT-4V}} &  &\textcolor{red!60}{\rule{0.5em}{0.5em}}0.00 	
    &\textcolor{red!60}{\rule{0.5em}{0.5em}}25.00 
    &\textcolor{red!60}{\rule{0.5em}{0.5em}}0.00 	
    &\textcolor{red!60}{\rule{0.5em}{0.5em}}15.50 	
    &\textcolor{red!60}{\rule{0.5em}{0.5em}}65.50 	
    &\textcolor{red!60}{\rule{0.5em}{0.5em}}60.00 	
    &\textcolor{red!60}{\rule{0.5em}{0.5em}}0.00 	
    &\textcolor{red!60}{\rule{0.5em}{0.5em}}20.00 	
    &\textcolor{red!60}{\rule{0.5em}{0.5em}}33.00 	
    &\textcolor{red!60}{\rule{0.5em}{0.5em}}0.00 	
    &\textcolor{red!60}{\rule{0.5em}{0.5em}}0.00 	
    &\textcolor{red!60}{\rule{0.5em}{0.5em}}0.00 	
    &\textcolor{red!60}{\rule{0.5em}{0.5em}}0.00 	
    &\textcolor{red!60}{\rule{0.5em}{0.5em}}0.00 	
    &\textcolor{red!60}{\rule{0.5em}{0.5em}}0.00 	
    &\textcolor{red!60}{\rule{0.5em}{0.5em}}76.00 	
    &\textcolor{red!60}{\rule{0.5em}{0.5em}}80.00 	
    &\textcolor{red!60}{\rule{0.5em}{0.5em}}0.00 	
    &\textcolor{red!60}{\rule{0.5em}{0.5em}}20.91  \\
      
    &&\textcolor{yellow!60}{\rule{0.5em}{0.5em}}0.00 	
    &\textcolor{yellow!60}{\rule{0.5em}{0.5em}}46.00 	
    &\textcolor{yellow!60}{\rule{0.5em}{0.5em}}2.50 	
    &\textcolor{yellow!60}{\rule{0.5em}{0.5em}}21.00 	
    &\textcolor{yellow!60}{\rule{0.5em}{0.5em}}65.50 	
    &\textcolor{yellow!60}{\rule{0.5em}{0.5em}}60.00 	
    &\textcolor{yellow!60}{\rule{0.5em}{0.5em}}0.00 	
    &\textcolor{yellow!60}{\rule{0.5em}{0.5em}}23.00 	
    &\textcolor{yellow!60}{\rule{0.5em}{0.5em}}33.00 	
    &\textcolor{yellow!60}{\rule{0.5em}{0.5em}}10.00 	
    &\textcolor{yellow!60}{\rule{0.5em}{0.5em}}0.00 	
    &\textcolor{yellow!60}{\rule{0.5em}{0.5em}}12.00 	
    &\textcolor{yellow!60}{\rule{0.5em}{0.5em}}3.00 	
    &\textcolor{yellow!60}{\rule{0.5em}{0.5em}}0.00 	
    &\textcolor{yellow!60}{\rule{0.5em}{0.5em}}20.00 	
    &\textcolor{yellow!60}{\rule{0.5em}{0.5em}}87.00 	
    &\textcolor{yellow!60}{\rule{0.5em}{0.5em}}80.00 	
    &\textcolor{yellow!60}{\rule{0.5em}{0.5em}}0.00 	
    &\textcolor{yellow!60}{\rule{0.5em}{0.5em}}26.00  \\

    && \textcolor{green!60}{\rule{0.5em}{0.5em}}0.00 	
    & \textcolor{green!60}{\rule{0.5em}{0.5em}}53.00 	
    & \textcolor{green!60}{\rule{0.5em}{0.5em}}24.50 	
    & \textcolor{green!60}{\rule{0.5em}{0.5em}}41.00 	
    & \textcolor{green!60}{\rule{0.5em}{0.5em}}67.00 	
    & \textcolor{green!60}{\rule{0.5em}{0.5em}}80.00 	
    & \textcolor{green!60}{\rule{0.5em}{0.5em}}9.00 	
    & \textcolor{green!60}{\rule{0.5em}{0.5em}}21.00 	
    & \textcolor{green!60}{\rule{0.5em}{0.5em}}49.00 	
    & \textcolor{green!60}{\rule{0.5em}{0.5em}}18.00 	
    & \textcolor{green!60}{\rule{0.5em}{0.5em}}0.00 	
    & \textcolor{green!60}{\rule{0.5em}{0.5em}}20.00 	
    & \textcolor{green!60}{\rule{0.5em}{0.5em}}22.00 	
    & \textcolor{green!60}{\rule{0.5em}{0.5em}}0.00 	
    & \textcolor{green!60}{\rule{0.5em}{0.5em}}62.00 	
    & \textcolor{green!60}{\rule{0.5em}{0.5em}}88.00 	
    & \textcolor{green!60}{\rule{0.5em}{0.5em}}80.00 	
    & \textcolor{green!60}{\rule{0.5em}{0.5em}}0.00 	
    & \textcolor{green!60}{\rule{0.5em}{0.5em}}36.09  \\
    \cmidrule(l){3-21}

    \multirow{3}*{\textbf{ChartVLM-B}} &  &\textcolor{red!60}{\rule{0.5em}{0.5em}}10.60 	
    &\textcolor{red!60}{\rule{0.5em}{0.5em}}20.40 	
    &\textcolor{red!60}{\rule{0.5em}{0.5em}}26.30 	
    &\textcolor{red!60}{\rule{0.5em}{0.5em}}29.10 	
    &\textcolor{red!60}{\rule{0.5em}{0.5em}}40.70 	
    &\textcolor{red!60}{\rule{0.5em}{0.5em}}15.80 	
    &\textcolor{red!60}{\rule{0.5em}{0.5em}}0.00 	
    &\textcolor{red!60}{\rule{0.5em}{0.5em}}38.00 	
    &\textcolor{red!60}{\rule{0.5em}{0.5em}}12.80 	
    &\textcolor{red!60}{\rule{0.5em}{0.5em}}0.00 	
    &\textcolor{red!60}{\rule{0.5em}{0.5em}}0.00 	
    &\textcolor{red!60}{\rule{0.5em}{0.5em}}0.00 	
    &\textcolor{red!60}{\rule{0.5em}{0.5em}}0.00 	
    &\textcolor{red!60}{\rule{0.5em}{0.5em}}0.00 	
    &\textcolor{red!60}{\rule{0.5em}{0.5em}}0.00 	
    &\textcolor{red!60}{\rule{0.5em}{0.5em}}28.08 	
    &\textcolor{red!60}{\rule{0.5em}{0.5em}}76.00 	
    &\textcolor{red!60}{\rule{0.5em}{0.5em}}0.00 	
    &\textcolor{red!60}{\rule{0.5em}{0.5em}}18.49  \\
      
    && \textcolor{yellow!60}{\rule{0.5em}{0.5em}}17.70 	
    & \textcolor{yellow!60}{\rule{0.5em}{0.5em}}27.50 	
    & \textcolor{yellow!60}{\rule{0.5em}{0.5em}}42.90 	
    & \textcolor{yellow!60}{\rule{0.5em}{0.5em}}45.00 	
    & \textcolor{yellow!60}{\rule{0.5em}{0.5em}}41.50 	
    & \textcolor{yellow!60}{\rule{0.5em}{0.5em}}15.80 	
    & \textcolor{yellow!60}{\rule{0.5em}{0.5em}}1.60 	
    & \textcolor{yellow!60}{\rule{0.5em}{0.5em}}67.00 	
    & \textcolor{yellow!60}{\rule{0.5em}{0.5em}}12.80 	
    & \textcolor{yellow!60}{\rule{0.5em}{0.5em}}5.80 	
    & \textcolor{yellow!60}{\rule{0.5em}{0.5em}}0.00 	
    & \textcolor{yellow!60}{\rule{0.5em}{0.5em}}2.20 	
    & \textcolor{yellow!60}{\rule{0.5em}{0.5em}}0.80 	
    & \textcolor{yellow!60}{\rule{0.5em}{0.5em}}0.00 	
    & \textcolor{yellow!60}{\rule{0.5em}{0.5em}}12.20 	
    & \textcolor{yellow!60}{\rule{0.5em}{0.5em}}33.46 	
    & \textcolor{yellow!60}{\rule{0.5em}{0.5em}}77.00 	
    & \textcolor{yellow!60}{\rule{0.5em}{0.5em}}20.40 	
    & \textcolor{yellow!60}{\rule{0.5em}{0.5em}}26.02  \\

    &&\textcolor{green!60}{\rule{0.5em}{0.5em}}21.20 	
    &\textcolor{green!60}{\rule{0.5em}{0.5em}}33.00 	
    &\textcolor{green!60}{\rule{0.5em}{0.5em}}51.90 	
    &\textcolor{green!60}{\rule{0.5em}{0.5em}}54.80 	
    &\textcolor{green!60}{\rule{0.5em}{0.5em}}43.20 	
    &\textcolor{green!60}{\rule{0.5em}{0.5em}}20.60 	
    &\textcolor{green!60}{\rule{0.5em}{0.5em}}13.20 	
    &\textcolor{green!60}{\rule{0.5em}{0.5em}}75.00 	
    &\textcolor{green!60}{\rule{0.5em}{0.5em}}15.20 	
    &\textcolor{green!60}{\rule{0.5em}{0.5em}}22.40 	
    &\textcolor{green!60}{\rule{0.5em}{0.5em}}4.20 	
    &\textcolor{green!60}{\rule{0.5em}{0.5em}}9.60 	
    &\textcolor{green!60}{\rule{0.5em}{0.5em}}1.60 	
    &\textcolor{green!60}{\rule{0.5em}{0.5em}}1.20 	
    &\textcolor{green!60}{\rule{0.5em}{0.5em}}18.40 	
    &\textcolor{green!60}{\rule{0.5em}{0.5em}}35.77 	
    &\textcolor{green!60}{\rule{0.5em}{0.5em}}77.80 	
    &\textcolor{green!60}{\rule{0.5em}{0.5em}}47.60 	
    &\textcolor{green!60}{\rule{0.5em}{0.5em}}32.65  \\
    \cmidrule(l){3-21}

    \multirow{3}*{\textbf{ChartVLM-L}} &  & \textcolor{red!60}{\rule{0.5em}{0.5em}}16.30 	
    & \textcolor{red!60}{\rule{0.5em}{0.5em}}34.00 	
    & \textcolor{red!60}{\rule{0.5em}{0.5em}}37.60 	
    & \textcolor{red!60}{\rule{0.5em}{0.5em}}34.70 	
    & \textcolor{red!60}{\rule{0.5em}{0.5em}}49.90 	
    & \textcolor{red!60}{\rule{0.5em}{0.5em}}24.80 	
    & \textcolor{red!60}{\rule{0.5em}{0.5em}}0.00 	
    & \textcolor{red!60}{\rule{0.5em}{0.5em}}45.80 	
    & \textcolor{red!60}{\rule{0.5em}{0.5em}}21.20 	
    & \textcolor{red!60}{\rule{0.5em}{0.5em}}0.00 	
    & \textcolor{red!60}{\rule{0.5em}{0.5em}}0.00 	
    & \textcolor{red!60}{\rule{0.5em}{0.5em}}0.00 	
    & \textcolor{red!60}{\rule{0.5em}{0.5em}}0.00 	
    & \textcolor{red!60}{\rule{0.5em}{0.5em}}0.00 	
    & \textcolor{red!60}{\rule{0.5em}{0.5em}}0.40 	
    & \textcolor{red!60}{\rule{0.5em}{0.5em}}23.65 	
    & \textcolor{red!60}{\rule{0.5em}{0.5em}}72.20 	
    & \textcolor{red!60}{\rule{0.5em}{0.5em}}0.00 	
    & \textcolor{red!60}{\rule{0.5em}{0.5em}}23.18  \\
      
    &&\textcolor{yellow!60}{\rule{0.5em}{0.5em}}19.50 	
    &\textcolor{yellow!60}{\rule{0.5em}{0.5em}}37.50 	
    &\textcolor{yellow!60}{\rule{0.5em}{0.5em}}55.80 	
    &\textcolor{yellow!60}{\rule{0.5em}{0.5em}}48.10 	
    &\textcolor{yellow!60}{\rule{0.5em}{0.5em}}49.90 	
    &\textcolor{yellow!60}{\rule{0.5em}{0.5em}}24.80 	
    &\textcolor{yellow!60}{\rule{0.5em}{0.5em}}0.40 	
    &\textcolor{yellow!60}{\rule{0.5em}{0.5em}}77.20 	
    &\textcolor{yellow!60}{\rule{0.5em}{0.5em}}21.20 	
    &\textcolor{yellow!60}{\rule{0.5em}{0.5em}}3.00 	
    &\textcolor{yellow!60}{\rule{0.5em}{0.5em}}1.80 	
    &\textcolor{yellow!60}{\rule{0.5em}{0.5em}}2.40 	
    &\textcolor{yellow!60}{\rule{0.5em}{0.5em}}2.00 	
    &\textcolor{yellow!60}{\rule{0.5em}{0.5em}}0.00 	
    &\textcolor{yellow!60}{\rule{0.5em}{0.5em}}19.60 	
    &\textcolor{yellow!60}{\rule{0.5em}{0.5em}}23.65 	
    &\textcolor{yellow!60}{\rule{0.5em}{0.5em}}72.20 	
    &\textcolor{yellow!60}{\rule{0.5em}{0.5em}}36.00 	
    &\textcolor{yellow!60}{\rule{0.5em}{0.5em}}30.68  \\

    &&\textcolor{green!60}{\rule{0.5em}{0.5em}}27.90 	
    &\textcolor{green!60}{\rule{0.5em}{0.5em}}42.00 	
    &\textcolor{green!60}{\rule{0.5em}{0.5em}}60.40 	
    &\textcolor{green!60}{\rule{0.5em}{0.5em}}55.10 	
    &\textcolor{green!60}{\rule{0.5em}{0.5em}}51.80 	
    &\textcolor{green!60}{\rule{0.5em}{0.5em}}28.40 	
    &\textcolor{green!60}{\rule{0.5em}{0.5em}}19.80 	
    &\textcolor{green!60}{\rule{0.5em}{0.5em}}87.20 	
    &\textcolor{green!60}{\rule{0.5em}{0.5em}}21.80 	
    &\textcolor{green!60}{\rule{0.5em}{0.5em}}27.60 	
    &\textcolor{green!60}{\rule{0.5em}{0.5em}}9.80 	
    &\textcolor{green!60}{\rule{0.5em}{0.5em}}8.00 	
    &\textcolor{green!60}{\rule{0.5em}{0.5em}}4.00 	
    &\textcolor{green!60}{\rule{0.5em}{0.5em}}0.60 	
    &\textcolor{green!60}{\rule{0.5em}{0.5em}}32.20 	
    &\textcolor{green!60}{\rule{0.5em}{0.5em}}25.19 	
    &\textcolor{green!60}{\rule{0.5em}{0.5em}}73.20 	
    &\textcolor{green!60}{\rule{0.5em}{0.5em}}69.20 	
    &\textcolor{green!60}{\rule{0.5em}{0.5em}}38.30  \\
    \hline
    
\end{tabular}
}

\label{tab:class-wise-se}
\end{table*}

\begin{table*}[t]\scriptsize
\renewcommand{\arraystretch}{1.6}
\centering
\caption{\textbf{Class-wise accuracy for Question Answering (QA) task evaluated using GPT-acc.}
}

\resizebox{\textwidth}{!}{%
\begin{tabular}{l|c|ccccc|ccc|ccc|cccc|ccc|c}
    \toprule
    &  & \multicolumn{5}{c|}{\textbf{General Chart Types}} & \multicolumn{13}{c|}{\textbf{Fine-grained Chart Types}} & \\
    \multirow{-2}{*}{Models} & \multirow{-2}{*}{Tasks} & bar & bar\_num & line & line\_num & pie &  ring & box & hist & treemap  &  rose  & area & 3D-bar &  bubble & multi & radar & heatmap & funnel & candle & \multirow{-2}{*}{Avg.} \\
    \hline
     
     % FILIP~\citep{yao2021filip} & $\times$ &
     % 89.8 & 99.2 & 99.8 & 75.0 & 93.4 & 96.3 & 61.3 & 84.3 & 90.4 & 45.9 & 70.6 & 79.3 & 82.1 \\
     \textbf{QWen-VL} & \multirow{9}*{QA} &33.00 &	31.00 &	22.00& 	22.00 &	45.00 &	24.00 &	16.00 &	24.00 &	20.00 &	10.00 &	10.00 &	16.00 &	16.00 &	8.00 &	16.00 &	26.92 &	28.00 &	14.00 &	23.26 \\
     \textbf{SPHINX-V2} &        &35.00 &51.00 &31.00 &25.00 &64.00 &30.00 &16.00 &30.00 &30.00 &22.00 &14.00 &18.00 &16.00 &12.00 &20.00 &40.38 &42.00 &14.00 &31.16 \\
     \textbf{ChartLlama} &        &14.00 &13.00 &9.00 &10.00 &39.00 &14.00 &20.00 &12.00 &18.00 &10.00 &8.00 &14.00 &12.00 &4.00 &16.00 &5.77 &10.00 &4.00 &13.80 \\
     \textbf{ChartAst} &        &36.00 &51.00 &29.00 &22.00 &67.00 &36.00 &14.00 &38.00 &28.00 &24.00 &14.00 &16.00 &14.00 &4.00 &22.00 &38.46 &40.00 &14.00 &30.99 \\
     \textbf{LLaVA-1.5} &        &24.00 &26.00 &10.00 &16.00 &29.00 &6.00 &30.00 &10.00 &22.00 &20.00 &12.00 &20.00 &22.00 &8.00 &18.00 &9.62 &8.00 &0.00 &17.18 \\
     \textbf{Matcha} &        &10.00 &18.00 &13.00 &12.00 &35.00 &6.00 &4.00 &10.00 &26.00 &10.00 &6.00 &8.00 &4.00 &4.00 &8.00 &19.23 &44.00 &6.00 &14.41 \\
     \textbf{GPT-4V} &        &20.00 &40.00 &25.00 &35.00 &65.00 &50.00 &30.00 &30.00 &70.00 &20.00 &10.00 &10.00 &30.00 &0.00 &30.00 &50.00 &60.00 &0.00 &33.04 \\
     \textbf{ChartVLM-B} &        &34.00 &38.00 &32.00 &37.00 &62.00 &44.00 &54.00 &40.00 &38.00 &16.00 &14.00 &16.00 &26.00 &18.00 &26.00 &40.38 &74.00 &26.00 &36.46 \\
     \textbf{ChartVLM-L} &        &41.00 &46.00 &33.00 &39.00 &68.00 &52.00 &56.00 &44.00 &44.00 &26.00 &26.00 &28.00 &24.00 &10.00 &24.00 &34.62 &80.00 &38.00 &40.71 \\
     \hline
\end{tabular}
}

\label{tab:class-wise-qa}
\end{table*}

\begin{table*}[t]\scriptsize
\renewcommand{\arraystretch}{1.6}
\centering

\caption{\textbf{Class-wise accuracy for Chart Description (Desc) evaluated using GPT-score.} The score of each individual description is an integer between 0-5.
}

\resizebox{\textwidth}{!}{%
\begin{tabular}{l|c|ccccc|ccc|ccc|cccc|ccc|c}
    \toprule
    &  & \multicolumn{5}{c|}{\textbf{General Chart Types}} & \multicolumn{13}{c|}{\textbf{Fine-grained Chart Types}} & \\
    \multirow{-2}{*}{Models} & \multirow{-2}{*}{Tasks} & bar & bar\_num & line & line\_num & pie &  ring & box & hist & treemap  &  rose  & area & 3D-bar &  bubble & multi & radar & heatmap & funnel & candle & \multirow{-2}{*}{Avg.} \\
    \hline
     
     % FILIP~\citep{yao2021filip} & $\times$ &
     % 89.8 & 99.2 & 99.8 & 75.0 & 93.4 & 96.3 & 61.3 & 84.3 & 90.4 & 45.9 & 70.6 & 79.3 & 82.1 \\
     \textbf{QWen-VL} & \multirow{8}*{Desc} &1.58 &1.30 &1.80 &1.75 &2.40 &1.60 &1.50 &1.70 &1.90 &1.50 &1.70 &1.50 &1.60 &1.80 &1.30 &1.60 &1.90 &1.30 &1.67 \\
     \textbf{SPHINX-V2} &        &1.36 &1.60 &1.50 &1.75 &2.35 &1.60 &1.00 &1.10 &1.70 &1.80 &1.30 &1.20 &1.40 &1.60 &1.30 &1.20 &1.80 &0.70 &1.53 \\
     \textbf{ChartLlama} &        &1.05 &1.00 &1.05 &1.00 &1.20 &1.10 &0.70 &1.10 &1.30 &1.20 &0.90 &0.90 &0.90 &1.20 &1.10 &1.50 &0.90 &0.60 &1.04 \\
     \textbf{ChartAst} &        &0.00 &0.40 &0.25 &0.15 &2.00 &0.90 &0.40 &0.00 &0.00 &0.60 &0.00 &0.00 &0.00 &0.00 &0.00 &0.00 &0.20 &0.00 &0.34 \\
     \textbf{LLaVA-1.5} &        &1.79 &1.30 &1.60 &1.70 &1.45 &1.10 &1.20 &1.20 &1.90 &2.00 &1.20 &1.80 &1.30 &1.60 &1.30 &1.60 &1.20 &1.10 &1.48 \\
     \textbf{GPT-4V} &        &2.84 &3.00 &2.95 &2.90 &3.55 &3.20 &3.10 &3.40 &3.60 &3.60 &3.40 &2.90 &3.50 &2.90 &3.00 &3.70 &3.70 &2.40 &3.17 \\
     \textbf{ChartVLM-B} &        &1.95 &2.70 &2.05 &1.90 &3.90 &2.40 &2.00 &2.40 &2.60 &1.60 &1.70 &1.70 &1.30 &1.50 &2.00 &2.60 &2.40 &2.40 &2.05 \\
     \textbf{ChartVLM-L} &        &1.47 &2.75 &2.45 &1.85 &4.00 &2.50 &2.60 &3.00 &2.50 &1.40 &0.90 &1.50 &1.00 &1.40 &1.00 &2.00 &3.30 &1.70 &2.17 \\
     \hline
\end{tabular}
}

\label{tab:class-wise-des}
\end{table*}

\begin{table*}[t]\scriptsize
\renewcommand{\arraystretch}{1.6}
\centering

\caption{\textbf{Class-wise accuracy for Chart Summarization (Summ) evaluated using GPT-score.} The score of each individual summarization is an integer between 0-5.
}

\resizebox{\textwidth}{!}{%
\begin{tabular}{l|c|ccccc|ccc|ccc|cccc|ccc|c}
    \toprule
    &  & \multicolumn{5}{c|}{\textbf{General Chart Types}} & \multicolumn{13}{c|}{\textbf{Fine-grained Chart Types}} & \\
    \multirow{-2}{*}{Models} & \multirow{-2}{*}{Tasks} & bar & bar\_num & line & line\_num & pie &  ring & box & hist & treemap  &  rose  & area & 3D-bar &  bubble & multi & radar & heatmap & funnel & candle & \multirow{-2}{*}{Avg.} \\
    \hline
     
     % FILIP~\citep{yao2021filip} & $\times$ &
     % 89.8 & 99.2 & 99.8 & 75.0 & 93.4 & 96.3 & 61.3 & 84.3 & 90.4 & 45.9 & 70.6 & 79.3 & 82.1 \\
     \textbf{QWen-VL} & \multirow{8}*{Summ} &1.58 &1.10 &1.55 &1.65 &1.95 &1.50 &1.40 &1.50 &1.60 &1.50 &1.50 &1.20 &1.20 &1.30 &1.40 &1.30 &1.50 &1.00 &1.45 \\
     \textbf{SPHINX-V2} &        &1.16 &1.60 &1.25 &1.10 &2.50 &1.40 &1.40 &1.50 &1.80 &1.40 &1.10 &1.10 &1.30 &1.10 &1.00 &1.10 &1.60 &1.10 &1.39 \\
     \textbf{ChartLlama} &        &1.05 &1.00 &0.95 &1.25 &1.00 &1.00 &1.00 &1.30 &1.10 &1.20 &0.80 &1.00 &0.70 &0.60 &1.30 &1.00 &0.70 &1.20 &1.02 \\
     \textbf{ChartAst} &        &1.00 &1.05 &0.85 &1.00 &2.40 &1.70 &2.70 &1.00 &0.30 &1.30 &0.50 &0.30 &1.60 &0.20 &0.70 &0.20 &0.30 &0.40 &1.03 \\
     \textbf{LLaVA-1.5} &        &1.42 &1.05 &2.00 &1.65 &1.30 &1.10 &1.10 &1.50 &1.30 &1.10 &1.00 &1.40 &1.30 &0.90 &1.20 &1.00 &0.80 &1.20 &1.29 \\
     \textbf{GPT-4V} &        &3.10 &2.80 &3.20 &2.75 &3.30 &3.10 &2.70 &4.00 &3.50 &3.60 &2.40 &2.70 &3.00 &3.10 &3.10 &4.10 &3.60 &2.70 &3.12 \\
     \textbf{ChartVLM-B} &        &1.26 &2.20 &1.95 &1.20 &3.30 &2.30 &2.70 &2.40 &2.40 &1.50 &1.00 &1.30 &1.00 &1.40 &1.00 &1.80 &2.30 &1.50 &1.84 \\
     \textbf{ChartVLM-L} &        &1.37 &2.50 &2.35 &1.90 &3.80 &3.00 &2.40 &2.90 &2.10 &1.30 &0.90 &1.00 &1.00 &1.40 &1.00 &1.70 &3.20 &1.30 &2.05 \\
     \hline
\end{tabular}
}

\label{tab:class-wise-summ}
\end{table*}

\begin{table*}[!t]\scriptsize
\renewcommand{\arraystretch}{1.6}
\centering
\caption{\textbf{Class-wise accuracy for Chart Re-drawing (Redraw) evaluated using GPT-score.} The score of each individual redrawing code is an integer between 0-5.
}

\resizebox{\textwidth}{!}{%
\begin{tabular}{l|c|ccccc|ccc|ccc|cccc|ccc|c}
    \toprule
    &  & \multicolumn{5}{c|}{\textbf{General Chart Types}} & \multicolumn{13}{c|}{\textbf{Fine-grained Chart Types}} & \\
    \multirow{-2}{*}{Models} & \multirow{-2}{*}{Tasks} & bar & bar\_num & line & line\_num & pie &  ring & box & hist & treemap  &  rose  & area & 3D-bar &  bubble & multi & radar & heatmap & funnel & candle & \multirow{-2}{*}{Avg.} \\
    \hline
     
     % FILIP~\citep{yao2021filip} & $\times$ &
     % 89.8 & 99.2 & 99.8 & 75.0 & 93.4 & 96.3 & 61.3 & 84.3 & 90.4 & 45.9 & 70.6 & 79.3 & 82.1 \\
      \textbf{QWen-VL} & \multirow{8}*{Redraw} &0.89 &0.60 &0.80 &1.30 &1.25 &1.10 &0.80 &0.80 &0.80 &1.10 &0.60 &1.10 &0.90 &0.60 &0.50 &0.50 &0.70 &0.70 &0.86 \\
     \textbf{SPHINX-V2} &        &1.00 &1.75 &1.60 &1.65 &1.80 &0.50 &0.40 &1.60 &0.60 &1.10 &0.20 &0.50 &0.40 &0.20 &0.00 &0.50 &0.30 &0.20 &0.96 \\
     \textbf{ChartLlama} &        &1.16 &1.05 &0.90 &1.15 &1.80 &0.70 &0.80 &1.00 &1.00 &0.70 &0.70 &1.10 &0.70 &0.40 &0.50 &1.00 &0.70 &0.30 &0.94 \\
     \textbf{ChartAst} &        &0.95 &1.35 &0.00 &0.60 &0.30 &0.00 &1.50 &2.40 &0.60 &1.70 &1.80 &0.60 &0.60 &1.20 &0.00 &0.00 &2.10 &0.00 &0.82 \\
     \textbf{LLaVA-1.5} &        &0.95 &0.75 &0.80 &0.95 &0.90 &0.60 &0.60 &0.80 &0.70 &1.00 &0.60 &0.80 &0.90 &0.40 &0.60 &0.70 &0.50 &0.50 &0.75 \\
     \textbf{GPT-4V} &        &2.05 &2.70 &2.05 &2.75 &3.55 &3.40 &2.00 &2.70 &2.70 &2.80 &2.20 &2.70 &2.40 &2.80 &2.30 &3.20 &3.50 &1.60 &2.63 \\
     \textbf{ChartVLM-B} &        &1.63 &1.50 &1.70 &1.65 &1.90 &1.10 &1.90 &1.10 &0.40 &1.20 &0.80 &1.00 &1.70 &1.30 &0.80 &1.20 &1.00 &1.10 &1.36 \\
     \textbf{ChartVLM-L} &        &1.53 &1.85 &1.85 &1.70 &2.75 &1.90 &1.40 &1.20 &0.90 &1.00 &1.10 &1.60 &1.30 &1.50 &0.80 &1.90 &1.20 &1.10 &1.58 \\
     \hline
\end{tabular}
}

\label{tab:class-wise-redraw}
\end{table*}

\begin{table}[t]
\renewcommand{\arraystretch}{1.4}
\centering
\caption{	
     Structural Extraction (SE) performance of baselines on real-world chart datasets (i.e., ChartQA and ChartX).
}
\scalebox{0.78}{
\begin{tabular}{l|l|l|l|l|l}
\toprule
\textbf{Model}     & \textbf{\#Params} & \textbf{val\_set} & \textbf{AP@Strict} & \textbf{AP@Slight} & \textbf{AP@High} \\ \hline
SPHINX-V2          & 13B               & ChartX            & 10.95              & 23.75              & 32.07            \\ 
ChartAst           & 13B               & ChartX            & 11.35              & 22.77              & 30.18            \\ 
Our ChartVLM-Base  & 7.3B              & ChartX            & 18.49              & 26.02              & 32.65            \\ 
Our ChartVLM-Large & 14.3B             & ChartX            & \textbf{23.18}              & \textbf{30.68}              &  \textbf{38.30}            \\  \hline
SPHINX-V2          & 13B               & ChartQA-Val       & 71.10              & 81.69              & 83.31            \\ 
ChartAst           & 13B               & ChartQA-Val       & 61.90              & 78.03              & 80.92            \\ 
Our ChartVLM-Base  & 7.3B              & ChartQA-Val       & 72.31              &  \textbf{84.24}              & 86.43            \\ 
Our ChartVLM-Large & 14.3B             & ChartQA-Val       & \textbf{73.13}              & 84.18              &  \textbf{87.10}            \\ \hline
\end{tabular}}
\vspace{6pt}

\label{tab:real_data}
\end{table}

\begin{table}[t]
\renewcommand{\arraystretch}{1.4}
\centering
\caption{Ablation study on the designed instruction adapter.}
\scalebox{0.70}{
\begin{tabular}{l|l|l|l|l|l}
\hline
\textbf{Input of auxiliary decoder} & \textbf{Model} & \textbf{QA: GPT-acc} & \textbf{Summ: GPT-score} & \textbf{Des: GPT-score} & \textbf{Redraw: GPT-score} \\ \hline
user instruction                    & ChartVLM-B     & 33.42                & 1.51                     & 1.67                    & 1.07                       \\ 
fixed task-specific instruction     & ChartVLM-B     & 36.4                 & 1.84                     & 2.05                    & 1.36                       \\ \hline
\end{tabular}}
\vspace{6pt}

\label{tab:abl_ins}
\end{table}

\subsection{Evaluation Settings}
\label{sec:eval_setting}
\noindent \textbf{Prompt Design for GPT-acc and GPT-score. }
We adopt GPT-acc as the evaluation metric for the QA task, and GPT-score for the description, summarization, and redrawing tasks, respectively. The complete prompts and manual criteria are concluded in Fig.~\ref{fig:gpt_eval_1} and~\ref{fig:gpt_eval_2}. 

\vspace{-2pt}
\begin{figure*}[!t]
    \centering
    \resizebox{0.95\linewidth}{!}{\includegraphics{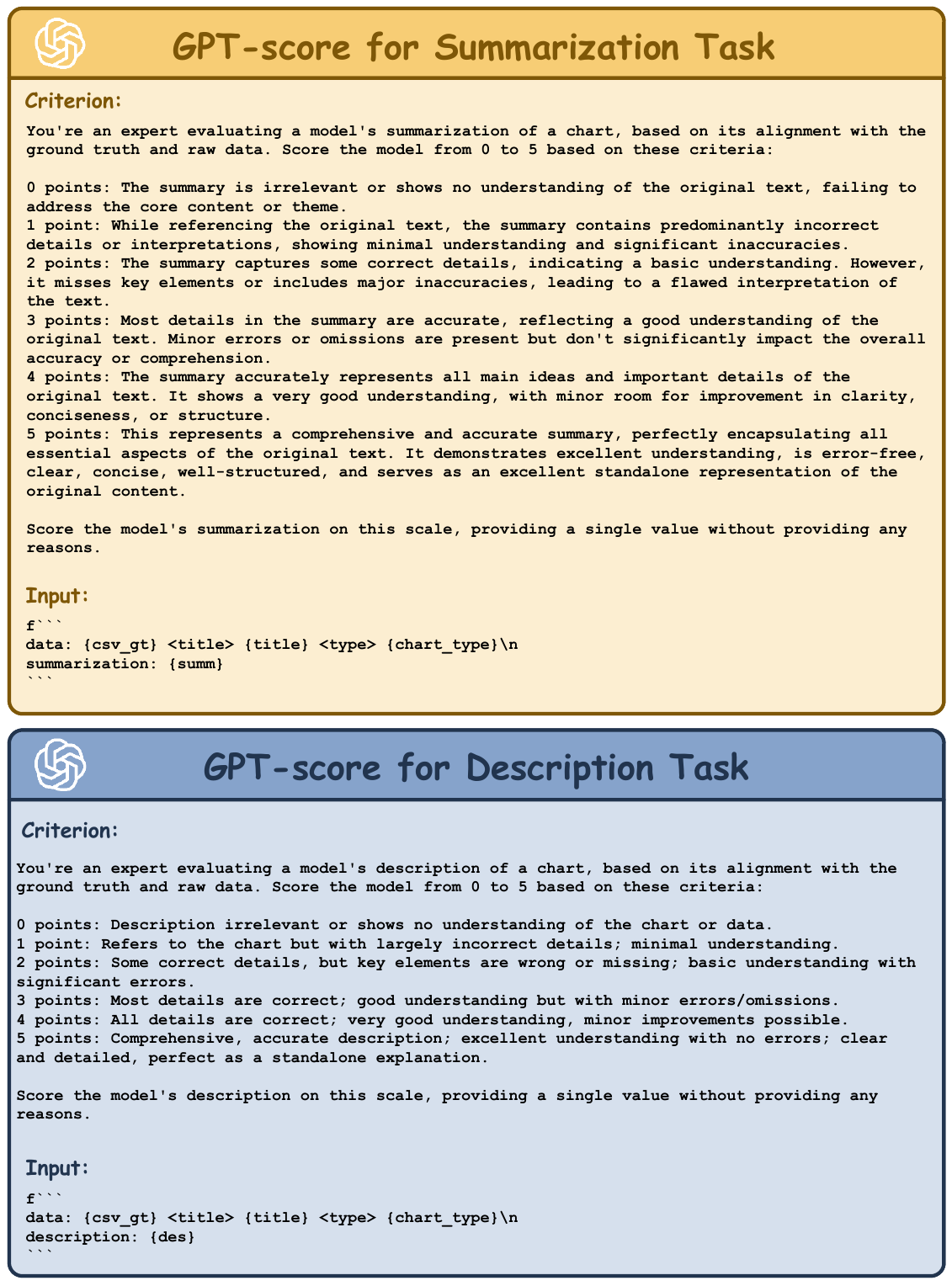}}
    % \vspace{-8pt}
    \caption{Detailed prompts in GPT-score metric for summarization and description tasks.} 
    \label{fig:gpt_eval_1}
    \vspace{-8pt}
\end{figure*}

\vspace{-2pt}
\begin{figure*}[!t]
    \centering
    \resizebox{1.05\linewidth}{!}{\includegraphics{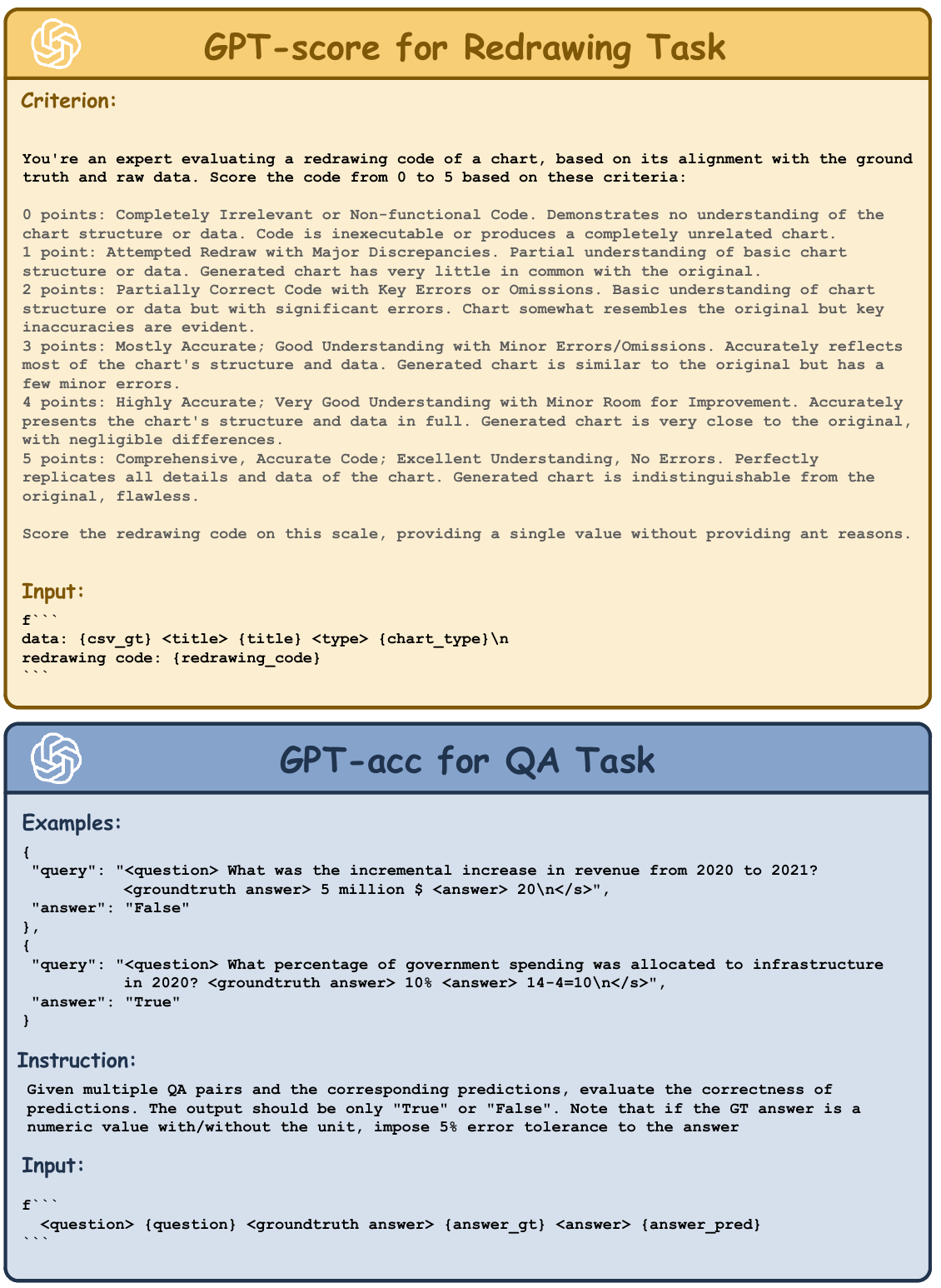}}
    % \vspace{-8pt}
    \caption{Detailed prompts in GPT-score metric for redrawing task and GPT-acc metric for QA task.} 
    \label{fig:gpt_eval_2}
    \vspace{-8pt}
\end{figure*}

\noindent \textbf{Employed Threshold of SCRM. }
\label{sec:scrm_tol}
According to the definition of SCRM metric proposed in StructChart~\cite{xia2023structchart}, three different levels of tolerance ( $tol:=\{ strict, slight, high\}$ ) are set for fine-grained evaluation of SE task. Considering the different perception difficulties of different types of charts, we divide all 18 types of charts into two difficulty levels: normal and difficult, and set different thresholds for tolerance respectively.

For normal charts, including \textit{bar chart, line chart, pie chart, bar chart with number, line chart with number, ring chart, heatmap, box plot, candlestick, funnel chart, histogram, and treemap}:
\begin{equation}
\small
\begin{aligned}
strict:=&\left\{J_{thr}|_{tol}=0   \wedge   e_{thr}|_{tol}=0\right\}, \\
slight:=&\left\{J_{thr}|_{tol}=2   \wedge  e_{thr}|_{tol}=0.05\right\}, \\
high:=&\left\{J_{thr}|_{tol}=5   \wedge  e_{thr}|_{tol}=0.1\right\}, \\
\end{aligned}
\end{equation}
For difficult charts, including \textit{rose chart, area chart, 3D-Bar chart, bubble chart, multi-axes chart, and radar chart}:
\begin{equation}
\small
\begin{aligned}
strict:=&\left\{J_{thr}|_{tol}=0   \wedge   e_{thr}|_{tol}=0.1\right\}, \\
slight:=&\left\{J_{thr}|_{tol}=2   \wedge  e_{thr}|_{tol}=0.3\right\}, \\
high:=&\left\{J_{thr}|_{tol}=5   \wedge  e_{thr}|_{tol}=0.5\right\}, \\
\end{aligned}
\end{equation}

\noindent where $J_{thr}|_{tol}$ indicates the edit distance threshold between prediction and GT string, $e_{thr}|_{tol}$ refers to the relative error threshold between prediction numeric value and GT value.

\subsection{Maximum number of generate token settings}

To fairly compare the performance of models on various tasks, we unify the maximum number of generate token (\texttt{max\_token}) of different models on the same task. The details of \texttt{max\_token} can be concluded: 1) 1280 for SE, 2) 100 for title, 3) 20 for type, 4) 100 for QA, 5) 512 for description and summarization, and 6) 1024 for redrawing code. This setting is still maintained for inference speed testing

\subsection{Quantitative Results for Each Chart Type}
\label{sec:each_class}
We have presented part of the class-wise performance in Fig. \ref{fig:class-wise} of the main text. Here, more comprehensive testing results of various models on all tasks are listed in Tables \ref{tab:class-wise-se},~\ref{tab:class-wise-qa},~\ref{tab:class-wise-des},~\ref{tab:class-wise-summ}, and~\ref{tab:class-wise-redraw}. Specifically, we compare recent multi-modal language models and chart-related models with ChartVLMs on QA, SE, description, summarization and redrawing tasks. The results show a comprehensive superiority of ChartVLMs to the existing models in most chart types and tasks. It should be noted that except for GPT-4V, whose scores of summary and description are higher than the average score, the downstream reasoning tasks seem quite tough for all models. This shed light on the common challenge in learning chart-related language models: how to fully learn multiple tasks in a single model without sacrificing the generalization ability to a new chart domain.

\subsection{Visualization Results of Perception Tasks}
\label{sec:vis_result_perception_tasks}
We provide four visualization perception results for different types of charts in Fig.~\ref{fig:perception_result}, including funnel chart, histogram, radar chart and line chart. The results demonstrate that our ChartVLM performs well on chart title and the chart-type prediction task. Even if the SE result of the radar chart is slightly wrong, ChartVLM still has strong SE performance on the funnel chart, histogram, and line chart. 

\vspace{-2pt}
\begin{figure*}[!t]
    \centering
    \resizebox{1.05\linewidth}{!}{\includegraphics{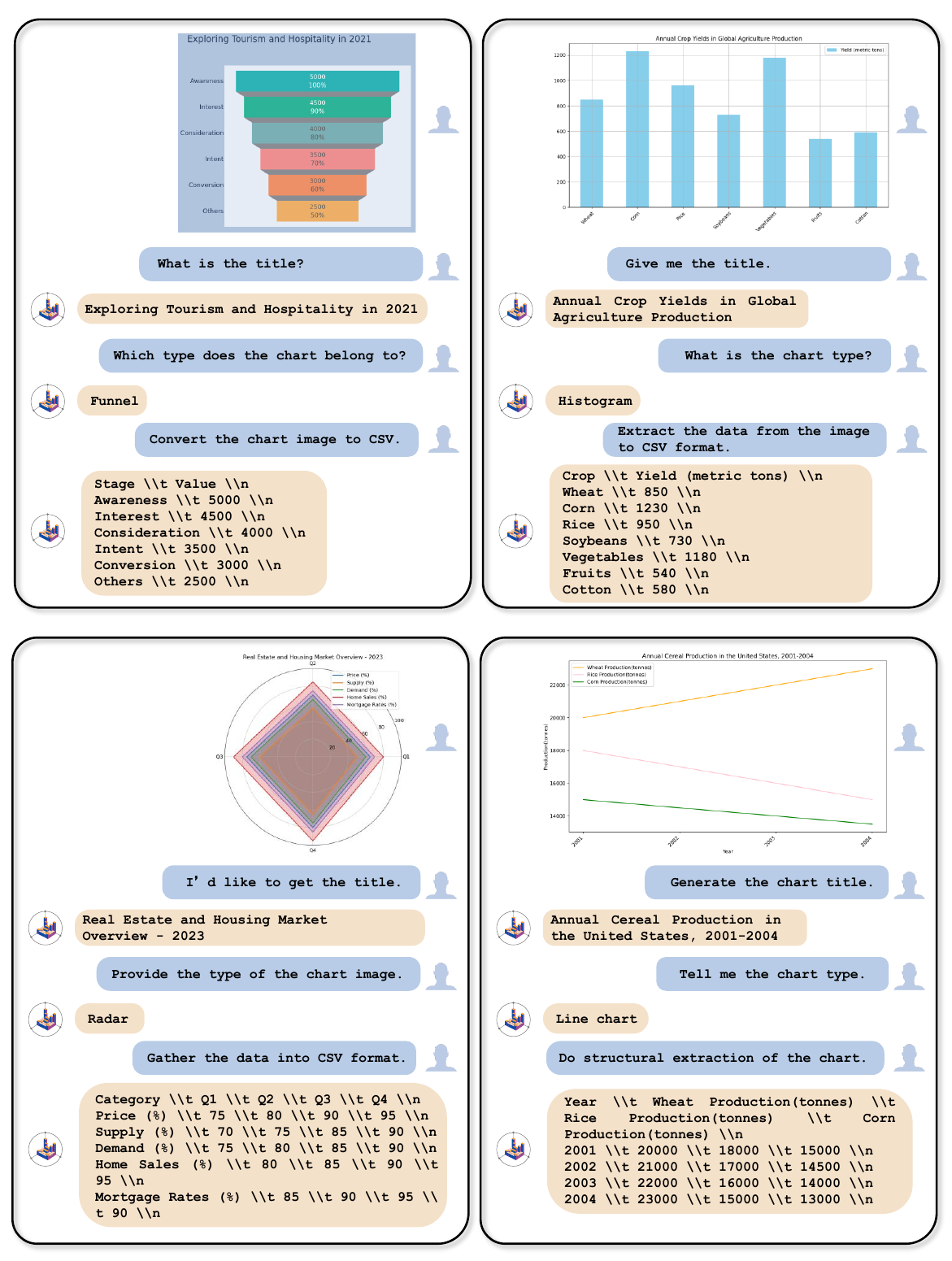}}
    % \vspace{-8pt}
    \caption{More visualization results for perception tasks using ChartVLM, including Structural Extraction (SE), chart title, and chart type prediction tasks.} 
    \label{fig:perception_result}
    \vspace{-8pt}
\end{figure*}

% \section{Details of ChartVLM Model}

% \subsection{Architecture of Image Encoder and Base Decoder}
% The image encoder and base decoder are based on ViT~\citep{Dosovitskiy2020AnII} structure. 

% While the bulk of the model is fairly standard, we propose a simple yet effective change. Instead of standard ViT, which scales the input images to a predefined resolution, we propose to always scale the input image to a fixed number of patches that can fit within the longest given sequence length, according to the original resolution of the input image. Moreover, we add 2-dimensional absolute positional embedding for the input patches, allowing the image encoder to handle variable resolutions.

\section{More Comprehensive Analysis}
\textbf{Comparison and Relationship between ChartX and Real-world Chart Datasets.} The controllable diversity of ChartX benefits the evaluation of both perception and cognition tasks more than the noise diversity of real-world chart datasets does. The chart data in ChartX are more diverse in the data representations than collected real-world charts, e.g., multiple modalities to represent each chart data and various chart types to represent different domains, etc. At current stages, real-world datasets are limited by high acquisition costs and annotation complexities, leading to biases in domain, data magnitude, and representation. They also lack scalability and have a narrow noise distribution. ChartX, however, offers varied data representations through multiple modalities and chart types. It uses controllable generalization tools and random prompt tuning to ensure diverse data distribution, aiding in a comprehensive evaluation of chart-related tasks at a lower cost. Unlike real-world charts, the synthetic charts in ChartX are clearer and more controllable, better suited for learning robust features in chart image understanding, thereby offering a strong foundation in diverse noise management.

The conclusions made on ChartX can transfer to existing real-world chart distributions. We provide the perception results of various MLLMs on both ChartX and ChartQA, representing synthetic and real-world chart distributions, respectively. As shown in Table~\ref{tab:real_data} , the better a model performs on ChartX, the better it performs on ChartQA, demonstrating that ChartX works as a comprehensive evaluation dataset where the conclusions can be well transferred to existing real-world chart distributions.

\textbf{Insights about Model Design.} Compared with end-to-end MLLMs, we design a cascade mechanism to decouple chart-related tasks into two stages: perception and cognition. Each stage features its own dedicated base decoder and auxiliary decoder, respectively. For percpetion tasks, only the lightweight decoder (base decoder) will participate in inference. For cognition tasks, both base decoder and a heavier one (auxiliary decoder) will participant. Some insights about the design of cascaded mechanism include:

\textbf{1)} The cascade mechanism can bring faster inference speed, especially for perception task, since only lightweight base decoder will be responsible for perception task. The experimental results in Table~\ref{tab:speed_diff} of the main text indicate the speed advantages brought by the cascade mechanism compared to other end-to-end MLLMs.
Chart represetation (title+type+csv) extracted by base decoder will provide interpretability for downstream cognition task. In Table~\ref{tab:abl_accumu} of the main text, the performance of cognition tasks will become better as the accuracy of chart representation increases (refer to SE performance).

\textbf{2)} The instruction adapter is trained to categorize user commands into 8 classes (7 task labels + 1 special label for tasks that fall outside the current capabilities of the model), which will cooperate with cascade mechanism to dynamically determine which decoders participate in the current task. Some insights about the design of cascaded mechanism include: The special label for tasks that fall outside the current capabilities of the model will serve as a safeguard against the occurrence of erroneous outputs or ``hallucination" issues. After determining the task, we can input fixed task-specific instructions to auxiliary decoder instead of directly using the user's natural language instructions as input. We have supplement experiment to demonstrate that fixed user instructions improve model performance in Table~\ref{tab:abl_ins}.

\textbf{Practical Applications.}
\textbf{1)} Applications in Specialized Fields: Enhanced data analysis and visualization will be a primary use of advanced ChartX in sectors such as business and academia. ChartX will facilitate better decision-making by elucidating data trends and interrelationships. In finance, ChartX will help develop a foundation model in examining market trends and company performances, shaping informed investment strategies. Similarly, in healthcare, ChartX will assist in the interpretation of medical imagery and the innovation of pharmaceuticals, supporting accurate diagnoses and tailored treatment plans.

\textbf{2)} Diverse Data Integration for Scientific Discovery: ChartX will be pivotal in synthesizing information from different data types, such as economic line graphs, geographic heatmaps, and military radar charts. By learning the real chart respresenation from ChartX, it will help bridge the gap between images and text, allowing for knowledge transfer across various disciplines. This integration will pave the way for interdisciplinary insights, spurring innovation by revealing previously untouched areas of research and potential breakthroughs.

\end{document}